\documentclass[letterpaper,journal]{IEEEtran}
\usepackage{amsmath,amsfonts}
\usepackage{array}
\usepackage{textcomp}
\usepackage{stfloats}
\usepackage{placeins}
\usepackage{url}
\usepackage{verbatim}
\usepackage{graphicx}
\usepackage{booktabs} 
\usepackage{siunitx}     
\usepackage{cite}
\usepackage{amssymb}
\usepackage{bm}
\usepackage{multirow}
\usepackage{color}
\usepackage{tabularx}

\usepackage{hyperref}
\begin{document}

\title{Intra-Agent Motion and Inter-Agent Relation Learning for Map-Free Trajectory Prediction}

\author{Wenyi Xiong, Jian Chen,Wenhua Chen, Ziheng Qi}
\maketitle

\begin{abstract}
Trajectory prediction is essential for safe planning and decision-making in autonomous driving. Map-free prediction improves deployment flexibility by removing the dependence on high-definition maps, but it also loses explicit road-topology and lane-connectivity priors. Consequently, the model must recover stable motion cues from short histories while resolving prospective interaction relations without lane-level guidance. To address these challenges, this paper proposes a map-free trajectory prediction framework centered on a novel Goal-Conditioned Expert-Routed Transformer (GCERT). A lightweight intra-agent front-end first decomposes projected historical motion features into a smooth trend and a local detail residual through a centered moving average. Component-specific residual projections transform the two signals before a causal temporal Transformer summarizes their recombined evolution. GCERT employs dual-level expert routing: edge-level relation experts generate interaction-specific messages, while node-level state experts adapt the transformation after message aggregation. Predicted motion goals update the node representations and construct prospective relative-geometry and orientation edge features before the final interaction layer, thereby conditioning attention, relation-specific message generation, and state transformation. GCERT therefore determines not only which neighboring agents should be attended to, but also what relation-specific information should be transmitted and how the aggregated node state should be transformed. The resulting representations are decoded into multimodal future trajectories and jointly optimized with trajectory-regression, mode-classification, and goal-prediction objectives. 
\end{abstract}

\begin{IEEEkeywords}
	Goal-conditioned expert-routed Transformer, inter-agent relation learning,
	intra-agent motion learning, map-free trajectory prediction,
	trend--detail motion decomposition.
\end{IEEEkeywords}

\section{Introduction}
\IEEEPARstart{T}{rajectory} prediction plays a fundamental role in autonomous driving, as accurately forecasting the future motions of surrounding traffic participants is essential for safe planning, risk assessment, and decision-making. Recent vectorized prediction methods, represented by VectorNet \cite{gao2020vectornet}, have substantially advanced this task by encoding traffic agents and scene elements as structured features. Graph-based approaches \cite{liang2020learning,jia2023hdgt} capture dependencies among heterogeneous scene entities, while attention-based methods \cite{xiong2025fine,gomez2023efficient,xin2025multi,liu2024laformer,liu2021multimodal,feng2023macformer} dynamically aggregate information from agents and contextual elements. Many of these methods rely on high-definition (HD) maps to provide lane geometry, road connectivity, and traffic-rule constraints, which offer strong priors for both future motion and agent interaction reasoning.

Despite their effectiveness, HD maps require considerable construction and maintenance efforts and may be unavailable, incomplete, or outdated in dynamically changing environments. Map-free trajectory prediction \cite{ren2025mlb,luo2025hsti,schmidt2022crat} therefore aims to forecast future motion using historical agent trajectories and their interactions without explicit map priors. This setting improves deployment flexibility, but also places greater demands on motion representation and interaction reasoning. In particular, removing road topology not only eliminates geometric constraints on individual trajectories, but also removes useful cues for distinguishing how neighboring agents may interact in the future.

\begin{figure*}[t]
	\centering
	\includegraphics[width=\textwidth]{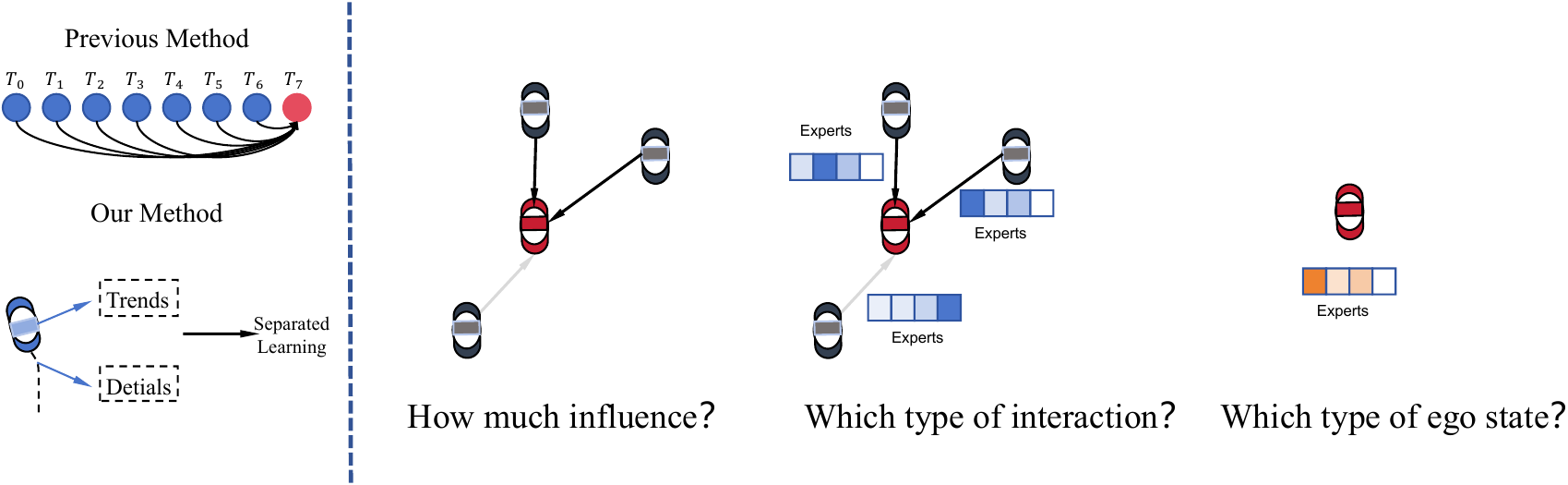}
	\caption{Motivation for the proposed framework. Left: conventional temporal encoding directly aggregates the observed sequence, whereas trend--detail decomposition separately represents persistent motion evolution and local deviations. Right: graph attention determines how much influence each neighbor contributes; edge-level relation experts determine which interaction-specific information is transmitted; and node-level state experts determine how the aggregated agent state is transformed.}
	\label{fig:motivation}
\end{figure*}

The motivation for our design is summarized in Fig.~\ref{fig:motivation}. On the intra-agent side, directly encoding the observed sequence may entangle persistent motion tendencies with local deviations. On the inter-agent side, conventional attention mainly determines how strongly each neighbor should contribute, but does not explicitly determine which relation-specific information should be transmitted or how the receiving node state should be transformed. Our framework addresses these three aspects through trend--detail motion decomposition, edge-level relation-expert routing, and node-level state-expert routing, respectively.

Map-free trajectory prediction therefore presents two related challenges:
\begin{itemize}
\item \textbf{Structured intra-agent motion representation.}
Historical trajectories simultaneously contain persistent motion tendencies and short-term deviations associated with maneuvers or local state changes. Directly encoding their superposition can obscure these complementary cues, particularly when explicit map constraints are absent. Separating them before temporal dependency modeling provides a more structured historical representation for subsequent interaction reasoning.
\item \textbf{Prospective inter-agent relation ambiguity.}
Without lane topology and road connectivity, agents with similar historical spatial configurations may develop substantially different future relations, such as following, merging, crossing, yielding, or separating. Existing interaction Transformers generally distinguish neighboring agents through attention weights, thereby determining how much information should be aggregated from each neighbor. However, heterogeneous interactions are commonly encoded within a shared value-transformation space. Such a formulation provides limited capability to determine what relation-specific information should be transmitted under different potential future relations, making inter-agent reasoning particularly ambiguous in map-free scenes.
\end{itemize}

To address these challenges, we propose a map-free trajectory prediction framework that combines a lightweight intra-agent motion front-end with prospective inter-agent relation reasoning. The front-end uses a centered moving-average operator to decompose projected historical motion features into trend and detail components. Independent residual projections transform the two components, and a causal temporal Transformer summarizes their recombined evolution.

We model inter-agent relations using GCERT, our proposed \emph{Goal-Conditioned Expert-Routed Transformer}. Unlike conventional interaction Transformers that encode heterogeneous relations through a shared value transformation, GCERT uses edge-level relation-expert routing to combine multiple relation-specific value transformations. A second, node-level router combines state experts after message aggregation. To compensate for the missing future-relation cues normally provided by map topology, GCERT predicts agent goals and constructs pairwise prospective relation features from their relative goal positions and motion orientations. Before the final interaction layer, the predicted-goal embedding updates each node representation, while the prospective pairwise features replace the history-only edge features. Consequently, the predicted goals jointly affect attention, relation routing, relation-specific message generation, and state-expert transformation. The resulting intra- and inter-agent representations are finally decoded into multimodal future trajectories.

Based on the above discussion, our contributions are summarized as follows:
\begin{itemize}
\item We establish a map-free trajectory prediction framework that jointly addresses structured intra-agent motion representation and prospective inter-agent relation ambiguity using only historical agent trajectories, achieving competitive prediction accuracy without HD maps.
\item We introduce a lightweight intra-agent front-end that separates projected histories into a smooth trend and a local detail residual, applies component-specific residual projections, and models their recombined temporal evolution with causal self-attention.
\item We introduce GCERT, a new interaction Transformer with dual-level expert routing. Relation experts determine what interaction-specific information is transmitted along each edge, while state experts adapt the node-level transformation after aggregation. Predicted goals update the node and edge representations before the final interaction layer, thereby conditioning attention, relation-specific message generation, and state-adaptive transformation.
\end{itemize}

\section{Related Work and Problem Formulation}
\subsection{Map-based and Map-free Trajectory Prediction}

Existing trajectory prediction methods can be broadly categorized into map-based and map-free approaches. Map-based methods exploit HD maps to encode lane geometry, road connectivity, and traffic-rule constraints. Representative methods such as LaneGCN \cite{liang2020learning}, HiVT \cite{zhou2022hivt}, HDGT \cite{jia2023hdgt}, and MTR \cite{shi2022motion} achieve strong performance through structured scene encoding and interaction reasoning. More recent approaches further incorporate large pretrained models or coupled map--agent representations \cite{lan2024traj,feng2023macformer}. Although explicit map priors improve the feasibility of predicted trajectories, their construction and maintenance costs restrict deployment in environments where maps are unavailable or rapidly changing.

Map-free methods instead infer future motion primarily from agent trajectories and interactions \cite{hou2024vehicle,gomez2023efficient,xiang2023map}. CRAT-Pred \cite{schmidt2022crat} models social context without relying on road geometry, MLB \cite{ren2025mlb} enhances agent-specific motion representations, and HSTI \cite{luo2025hsti} develops hierarchical spatial--temporal interaction modeling. MFIG-Net \cite{wang2025mfig} additionally exploits traffic-light observations as lightweight scene context. These studies demonstrate the feasibility of map-free prediction, but the absence of explicit topology leaves two fundamental sources of uncertainty: historical motion must provide sufficient information to characterize each agent, and inter-agent relations must be inferred without lane-level guidance.

\subsection{Intra-Agent Motion Representation}

Intra-agent motion representation aims to encode the temporal evolution of each agent from its historical observations. Recurrent architectures, including GRU- and LSTM-based models \cite{chung2014empirical,hochreiter1997long,xu2023context}, capture sequential dependencies through hidden-state propagation. Transformer-based predictors \cite{vaswani2017attention,liu2024laformer} provide a global temporal receptive field, while Mamba-based approaches \cite{gu2024mamba,ren2025mlb} offer an efficient alternative for long-sequence modeling. Nevertheless, directly processing the original sequence may entangle smooth motion tendencies with short-lived deviations.

Structured and multi-scale representations have therefore been explored to expose complementary temporal patterns \cite{bian2024multi,chen2024pathformer,zhang2023flight}. Following the general principle of separating slowly varying evolution from local deviations, our intra-agent front-end applies a centered moving average in the projected feature space to obtain a trend and a detail residual. Component-specific residual projections transform the two signals before temporal self-attention. This lightweight design preserves the original temporal resolution while providing a more structured input to the interaction model.

\subsection{Inter-Agent Relation Modeling}

Inter-agent interaction is commonly modeled using graphs or attention. Graph-based methods propagate messages along agent edges \cite{schmidt2022crat,zhang2024g2ltraj}, whereas Transformer-based methods dynamically weight social dependencies among agents. AgentFormer \cite{yuan2021agentformer} jointly models temporal and social dimensions, AutoBots \cite{girgis2021latent} performs set-based multi-agent prediction, and HiVT \cite{zhou2022hivt} applies hierarchical interaction learning. Goal-conditioned methods further use destinations or goal-related features to guide prediction and interaction aggregation \cite{li2024efficient,gulzar2025gc}. However, these methods generally represent heterogeneous interactions using shared feature transformations, even though the same neighboring agent may convey different information under following, merging, crossing, or yielding relations.

Mixture-of-experts models use input-dependent routing to combine specialized transformations \cite{shazeer2017outrageously}. Motivated by this principle, GCERT employs two complementary routing stages. The edge-level router infers a soft distribution over latent interaction relations and combines relation-specific value transformations, while the node-level router combines state experts after message aggregation. The purpose is not generic capacity expansion: relation-expert weights describe pairwise relation uncertainty, and state-expert weights adapt the transformation of the aggregated node representation. Unlike conventional goal conditioning that primarily guides trajectory decoding or attention, predicted goals update both the node and pairwise edge representations entering the final interaction layer and therefore influence its attention, relation-expert routing, and state-expert transformation.

\subsection{Problem Formulation}
Consider a traffic scene containing \(N\) agents. The historical observations are denoted by \(\mathbf{X}_{h}=\{\mathbf{p}_{i}^{t}\}\), where \(\mathbf{p}_{i}^{t}\in\mathbb{R}^{2}\) is the position of agent \(i\) at time \(t\), \(i=1,\ldots,N\), and \(t=-T_h+1,\ldots,0\). Given \(T_h\) historical steps, the objective is to predict \(K\) possible future trajectories and their confidence scores over a horizon of \(T_f\):
\begin{equation}
\label{eq:problem-formulation}
\left\{
\left(\hat{\mathbf{Y}}_{i}^{k},\pi_{i}^{k}\right)
\right\}_{k=1}^{K}
=
\mathcal{F}_{\theta}\left(\mathbf{X}_{h}\right),
\qquad
\sum_{k=1}^{K}\pi_i^k=1,
\end{equation}
where \(\hat{\mathbf{Y}}_{i}^{k}=\{\hat{\mathbf{p}}_{i,k}^{t}\}_{t=1}^{T_f}\) denotes the \(k\)-th future trajectory hypothesis of agent \(i\), \(\pi_i^k\) is its confidence, and \(\mathcal{F}_{\theta}\) represents the proposed prediction model. Benchmark evaluation is performed on the designated target agent.
For local prediction and supervision, a scene-frame position is expressed relative to the current pose of agent \(i\) as \(\mathbf y_i^t=\mathbf R_i^\top(\mathbf p_i^t-\mathbf p_i^0)\), where \(\mathbf R_i\) rotates the agent's heading-aligned local frame into the scene frame.

\section{Proposed Model}

\subsection{Map-free Trajectory Prediction Framework Overview}

\begin{figure*}
	\centering
	\includegraphics[width=\textwidth]{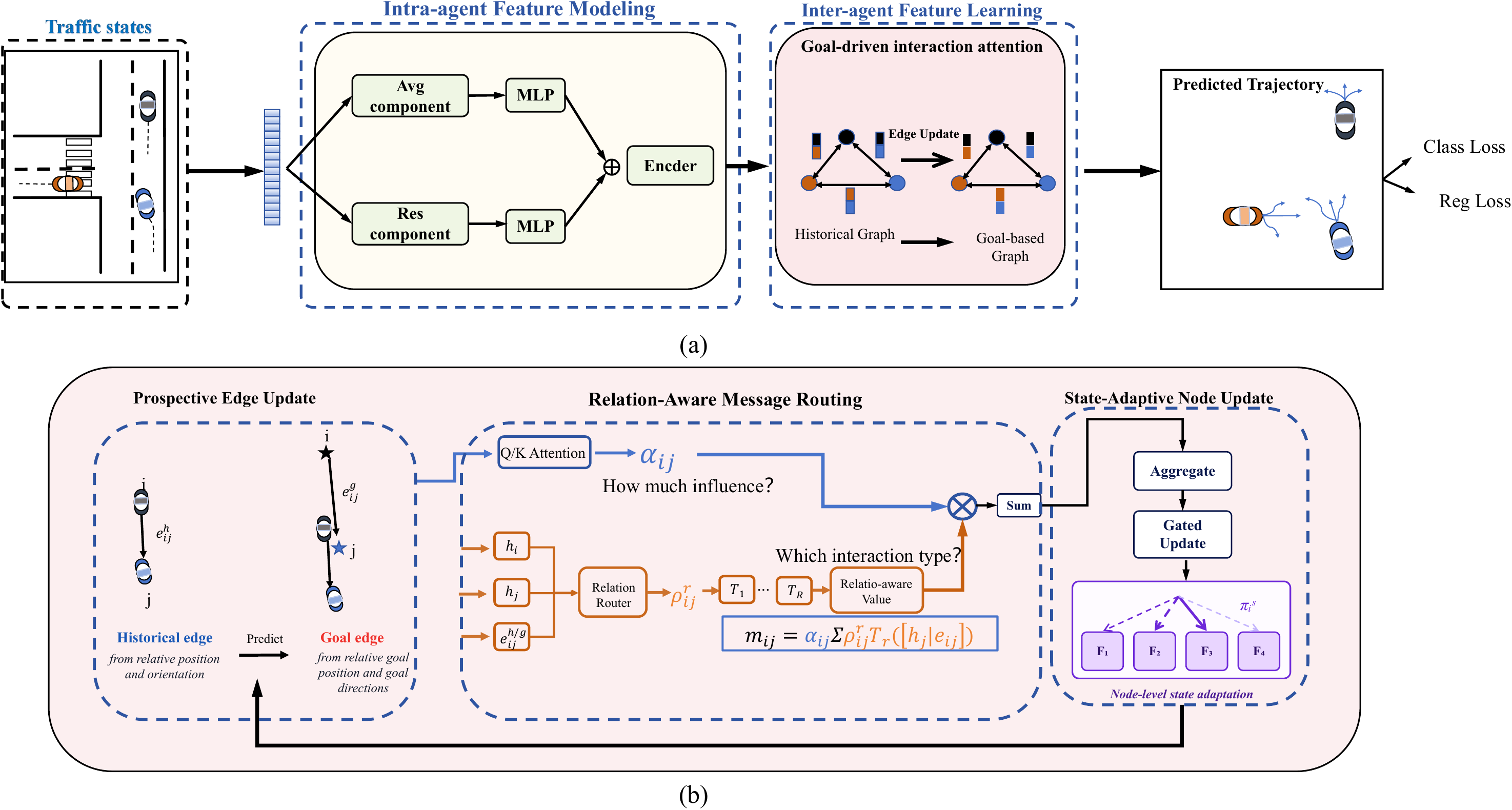}
\caption{
Overview of the proposed framework. Historical trajectories are encoded using trend--detail decomposition, component-specific residual projection, and causal temporal self-attention. GCERT performs goal-conditioned dual-level expert routing through edge-level relation experts and node-level state experts. The fused motion and interaction representations are decoded into multimodal future trajectories.
}
	\label{fig:frame}
\end{figure*}

Fig. \ref{fig:frame} presents the proposed map-free trajectory prediction framework, which consists of a lightweight intra-agent motion front-end, a Goal-Conditioned Expert-Routed Transformer (GCERT), and a multimodal trajectory decoder. Given historical agent trajectories, the intra-agent front-end decomposes projected motion features into trend and detail components, transforms and recombines them, and models their temporal dependencies with causal self-attention. GCERT then performs inter-agent reasoning through graph attention, edge-level relation-expert routing, and node-level state-expert routing. To reduce the prospective relation ambiguity caused by missing map topology, an intermediate goal prediction conditions the final interaction layer. The resulting local motion and global interaction representations are decoded into multiple future trajectories and their confidence scores.

\subsection{Intra-Agent Motion Modeling}

In map-free trajectory prediction, the historical trajectory is the primary source for inferring an agent's motion state and potential intention. Unlike map-based methods, the model cannot use lane curvature, lane direction, or road connectivity to compensate for an incomplete motion representation. A short history may simultaneously contain a persistent motion tendency and local deviations caused by acceleration, turning, or behavioral transitions. Directly encoding their superposition can make these complementary cues difficult to distinguish. We therefore use a lightweight front-end to decompose the projected history into a smooth trend and a detail residual, transform the two components independently, and model their recombined temporal evolution.

\subsubsection{Trend--Detail Motion Decomposition}

For agent \(i\), let \(\Delta\mathbf{p}_i^t\) denote the frame-wise displacement supplied by the preprocessed input. It is first projected into a \(D\)-dimensional latent space:
\begin{equation}
\label{eq:motion-projection}
\mathbf{E}_{i}^{t}
=
\mathbf{W}_{e}\Delta\mathbf{p}_{i}^{t}+\mathbf{b}_{e},
\qquad
\mathbf{E}_{i}\in\mathbb{R}^{T_h\times D}.
\end{equation}
The projection in Eq.~\eqref{eq:motion-projection} uses frame-wise displacements rather than absolute scene coordinates because they describe local motion changes and reduce sensitivity to the global scene origin. Decomposition is performed after projection so that every latent channel can represent a different response to the observed motion.

Let \(k\) be an odd smoothing-kernel length and \(r=(k-1)/2\). To preserve the sequence length, the first and last feature vectors are replicated \(r\) times before applying a centered moving average. The trend component is
\begin{equation}
\label{eq:trend-component}
\mathbf{M}_{i}^{t}
=
\frac{1}{k}
\sum_{\tau=-r}^{r}
\bar{\mathbf{E}}_{i}^{t+\tau},
\end{equation}
where \(\bar{\mathbf{E}}_{i}\) denotes the endpoint-replicated sequence. The detail component is defined as the residual
\begin{equation}
\label{eq:detail-component}
\mathbf{R}_{i}^{t}
=
\mathbf{E}_{i}^{t}-\mathbf{M}_{i}^{t}.
\end{equation}
The moving average in Eq.~\eqref{eq:trend-component} captures slowly varying motion evolution, whereas the residual in Eq.~\eqref{eq:detail-component} retains short-term deviations from that trend. This decomposition is deterministic and introduces no additional supervision.

The two components are transformed by independent component-specific residual projections. For \(c\in\{M,R\}\),
\begin{equation}
\label{eq:component-projection}
\mathbf{H}_{i}^{c}
=
\operatorname{LN}_{c}
\left(
\mathbf{Q}_{i}^{c}
+
\operatorname{Dropout}
\left(
\operatorname{GELU}
\left(
\mathbf{W}_{c}\mathbf{Q}_{i}^{c}+\mathbf{b}_{c}
\right)
\right)
\right),
\end{equation}
where \(\mathbf{Q}_{i}^{M}=\mathbf{M}_{i}\) and \(\mathbf{Q}_{i}^{R}=\mathbf{R}_{i}\); the linear maps act independently at each time step. The outputs of the component projections in Eq.~\eqref{eq:component-projection} are added elementwise:
\begin{equation}
\label{eq:component-fusion}
\tilde{\mathbf{E}}_{i}
=
\mathbf{H}_{i}^{M}+\mathbf{H}_{i}^{R}.
\end{equation}
Independent parameters allow the trend and detail components to be mapped differently, while the fusion in Eq.~\eqref{eq:component-fusion} restores a single feature sequence with the original temporal length. The decomposition is intended to structure the input rather than discard either component: the trend preserves persistent evolution, whereas the residual retains maneuver-related local changes. Finally, the fused sequence \(\tilde{\mathbf{E}}_{i}\) is encoded by a standard temporal Transformer with learnable positional embeddings, a causal attention mask, and a classification token. The trend--detail decomposition is applied before invalid historical steps are replaced by a learned padding token in the temporal Transformer. Fully observed histories contain no missing-step padding, although the first displacement is initialized to zero. For partially observed neighbors, missing-step displacements are set to zero before projection and participate in the moving average; their projected features can therefore influence valid steps within the smoothing radius ($r=6$). Replacing invalid steps with padding tokens afterward does not remove this boundary effect. The classification-token output summarizes the encoded history and forms the intra-agent representation \(\mathbf{h}_{i}\) used by the subsequent interaction model.

\subsection{Goal-Conditioned Expert-Routed Transformer}

The intra-agent features describe how individual agents have moved, but future prediction additionally requires reasoning about how their behaviors influence one another. This problem is especially challenging without map topology: spatially adjacent agents do not necessarily have the same behavioral relation, and similar historical configurations may lead to following, merging, crossing, yielding, or separating behaviors. A conventional attention layer can vary the scalar importance assigned to each neighbor, but its shared value transformation still encodes different relations in a common feature space. Motivated by Fig.~\ref{fig:motivation}, we develop GCERT to jointly model neighbor relevance, latent interaction relations, relation-specific information transmission, and state-adaptive node transformation. GCERT comprises prospective edge construction, relation-expert message routing, a gated node update, and state-expert routing.

The agents valid at the current time are organized as a directed interaction graph.
For an edge \(j\rightarrow i\), the history-based edge feature is constructed from the current relative position and heading difference:
\begin{equation}
\label{eq:historical-edge}
\begin{aligned}
\Delta\theta_{ij}&=\theta_j^0-\theta_i^0,\\
\mathbf{e}_{ij}&=\phi_e\left(
\mathbf{R}_i^{\mathsf{T}}(\mathbf{p}_j^0-\mathbf{p}_i^0)
,\,[\cos\Delta\theta_{ij},\sin\Delta\theta_{ij}]^{\mathsf T}
\right).
\end{aligned}
\end{equation}
In Eq.~\eqref{eq:historical-edge}, \(\theta_i^0\) is the current heading of agent \(i\), and \(\mathbf{R}_i\) maps local-frame column vectors to the scene frame. The edge encoder \(\phi_e\) embeds relative position and heading difference through separate MLPs, sums their embeddings, and applies normalization, a nonlinear activation, and a linear projection followed by normalization to produce a \(D\)-dimensional edge feature. This directed construction preserves the asymmetry of traffic interactions: the effect of agent \(j\) on agent \(i\) need not equal the reverse effect. Let \(\check{\mathbf h}_i=\operatorname{LN}_1(\mathbf h_i)\) denote the input to the interaction sublayer. Based on the target feature, neighboring feature, and edge geometry, a graph attention score is computed by
\begin{equation}
\label{eq:graph-attention}
\alpha_{ij}
=
\operatorname{SoftMax}_{j}
\left(
\frac{
\mathbf{q}_{i}^{\mathsf{T}}
\left(\mathbf{k}_{j}+\mathbf{k}_{ij}^{e}\right)
}{\sqrt{d}}
\right),
\end{equation}
where \(\mathbf{q}_{i}=\mathbf{W}_{q}\check{\mathbf{h}}_{i}\), \(\mathbf{k}_{j}=\mathbf{W}_{k}\check{\mathbf{h}}_{j}\), and \(\mathbf{k}_{ij}^{e}=\mathbf{W}_{ke}\mathbf{e}_{ij}\). The equation is written for one attention head, with the head index omitted for brevity and \(d=D/H\), where \(D\) is the embedding dimension and \(H\) is the number of heads. Queries and keys are computed from pre-normalized node features, and dropout is applied to the normalized attention weights during training. The edge-conditioned key allows relative geometry to affect the attention distribution. Nevertheless, \(\alpha_{ij}\) mainly determines how much information is received from neighbor \(j\). GCERT further introduces relation-expert routing to determine what interaction information should be transmitted.

\subsubsection{Relation-Conditioned Value Transformation}

Traffic relations are usually not explicitly annotated and may not fall into a single deterministic category. GCERT therefore treats them as latent variables and learns them jointly with trajectory prediction. For each directed edge, a relation router produces a soft distribution over \(R\) latent interaction relations:
\begin{equation}
\label{eq:relation-router}
\boldsymbol{\rho}_{ij}
=
\operatorname{SoftMax}
\left(
f_{\mathrm{rel}}
\left[
\check{\mathbf{h}}_{i}
\mathbin{\Vert}\check{\mathbf{h}}_{j}
\mathbin{\Vert}\mathbf{e}_{ij}
\right]
\right).
\end{equation}
The estimator considers the states of both agents together with their pairwise geometry. This enables two edges connected to the same neighboring agent to receive different relation assignments. In parallel, a bank of \(R\) relation-specific transformations maps the neighboring feature and its edge context into distinct values:
\begin{equation}
\label{eq:relation-expert}
\mathbf{v}_{ij}^{r}
=
\mathcal{T}_{r}
\left(
\left[
\check{\mathbf{h}}_{j}
\mathbin{\Vert}\mathbf{e}_{ij}
\right]
\right),
\qquad r=1,\ldots,R.
\end{equation}
Combining Eqs.~\eqref{eq:graph-attention}--\eqref{eq:relation-expert}, the message from agent \(j\) to agent \(i\) is formulated as
\begin{equation}
\label{eq:relation-message}
\mathbf{m}_{ij}
=
\alpha_{ij}
\sum_{r=1}^{R}
\rho_{ij}^{r}\mathbf{v}_{ij}^{r}.
\end{equation}
The relation-weighted message in Eq.~\eqref{eq:relation-message} is implemented with multiple heads: the fused relation value is split into \(H\) channel groups, weighted by the corresponding head-specific attention coefficients, and concatenated to form \(\mathbf{m}_{ij}\). Relation-routing probabilities are shared across heads. Unlike a shared value projection, this weighted relation transformation provides multiple relation-conditioned representation spaces. Each transformation can capture a different interaction response, while the soft relation distribution retains uncertainty when an edge cannot be assigned confidently to one relation. Consequently, the same neighboring feature can generate different messages for different target agents or traffic configurations. The attention coefficient \(\alpha_{ij}\) and relation probability \(\rho_{ij}^{r}\) play complementary roles: the former models neighbor relevance, whereas the latter controls the semantic form of the transmitted value.

The incoming messages are summed and combined with the target feature through a gated update:
\begin{equation}
\label{eq:gated-update}
\begin{aligned}
\mathbf{u}_{i}&=\sum_{j\in\mathcal{N}(i)}\mathbf{m}_{ij},\\
\mathbf{z}_{i}&=\operatorname{Sigmoid}
\left(\mathbf{W}_{u}\mathbf{u}_{i}+\mathbf{W}_{h}\check{\mathbf{h}}_{i}\right),\\
\mathbf{h}_{i}'&=\mathbf{u}_{i}
+\mathbf{z}_{i}\odot
\left(\mathbf{W}_{s}\check{\mathbf{h}}_{i}-\mathbf{u}_{i}\right).
\end{aligned}
\end{equation}
Bias terms are omitted from Eq.~\eqref{eq:gated-update}. The interaction residual is \(\mathbf a_i=\mathbf h_i+\operatorname{Dropout}(\mathbf W_o\mathbf h_i'+\mathbf b_o)\). Layer normalization precedes both the interaction and feed-forward sublayers; \(\bar{\mathbf{h}}_i\) denotes the output of the second normalization, given by \(\bar{\mathbf h}_i=\operatorname{LN}_2(\mathbf a_i)\). The conventional shared feed-forward sublayer is replaced by a dense soft mixture of \(S\) state experts. Given the normalized node state \(\bar{\mathbf{h}}_i\), the router and state transformation are
\begin{equation}
\label{eq:state-experts}
\begin{aligned}
\boldsymbol{\gamma}_{i}
&=\operatorname{SoftMax}\left(\mathbf{W}_{r}\bar{\mathbf{h}}_{i}+\mathbf{b}_{r}\right),\\
\mathbf{o}_{i}
&=\sum_{s=1}^{S}\gamma_i^s f_s\left(\bar{\mathbf{h}}_{i}\right).
\end{aligned}
\end{equation}
In Eq.~\eqref{eq:state-experts}, every expert is a two-layer feed-forward network and all experts are evaluated before soft weighted fusion. The complete layer output is \(\mathbf h_i^{(l)}=\mathbf a_i+\mathbf o_i\). Unlike the edge-wise relation experts, which determine the semantic content of each incoming message, these state experts adapt the node-level feature transformation after message aggregation. Stacking the resulting GCERT layers progressively updates agent features and allows relation-aware information to propagate across the interaction graph.

\subsubsection{Goal-Conditioned Prospective Relation Reasoning}

Although relation-specific transformations increase the diversity of interaction messages, relations inferred only from historical motion and current geometry may remain ambiguous. In map-based prediction, lane connectivity indicates whether two agents are likely to converge or separate. To provide an analogous prospective cue without maps, the first \(L-1\) GCERT layers perform history-based interaction reasoning. Let \(\mathbf{h}_i^{(l)}\) denote the complete output of GCERT layer \(l\), including both sublayers and their residual connections. The output of layer \(L-1\) is used to predict an auxiliary motion goal:
\begin{equation}
\label{eq:goal-prediction}
\hat{\mathbf{g}}_{i}=f_{g}\left(\mathbf{h}_{i}^{(L-1)}\right).
\end{equation}
The goal in Eq.~\eqref{eq:goal-prediction} is expressed as an offset from the current position and provides a compact estimate of future motion intention. Let \(\hat{\mathbf{s}}_i\) denote the resulting prospective endpoint and \(\hat{\mathbf{d}}_i\) its normalized goal direction in the scene coordinate system. Because \(\hat{\mathbf{g}}_i\) is predicted in agent \(i\)'s local frame, it is converted to the scene frame and added to the current position as
\begin{equation}
\label{eq:prospective-endpoint}
\hat{\mathbf{s}}_i=\mathbf{p}_i^0+\mathbf{R}_i\hat{\mathbf{g}}_i.
\end{equation}
For the endpoint in Eq.~\eqref{eq:prospective-endpoint}, when the goal-offset norm is at most \(10^{-6}\), the direction falls back to \((\cos\theta_i^0,\sin\theta_i^0)\), using the current heading angle supplied by the input data. For edge \(j\rightarrow i\), we construct a goal-conditioned relation feature using relative endpoint geometry and directional consistency:
\begin{equation}
\label{eq:goal-edge}
\mathbf{e}_{ij}^{g}
=
\phi_{g}
\left(
\mathbf{R}_{i}^{\mathsf{T}}(\hat{\mathbf{s}}_{j}-\hat{\mathbf{s}}_{i})
,\,\left[
\left(\hat{\mathbf{d}}_{j}^{\mathsf{T}}\hat{\mathbf{d}}_{i}\right)
,\,
\operatorname{cross}
\left(\hat{\mathbf{d}}_{i},\hat{\mathbf{d}}_{j}\right)
\right]^{\mathsf T}\right).
\end{equation}
In Eq.~\eqref{eq:goal-edge}, \(\mathbf{R}_i\) maps column vectors from agent \(i\)\textquotesingle s local frame to the scene frame, so its transpose expresses the relative endpoint in the receiving agent\textquotesingle s local frame. Here \(\operatorname{cross}(\mathbf a,\mathbf b)=a_xb_y-a_yb_x\) is a scalar. The dot product and planar cross product equal \(\cos(\theta_j^g-\theta_i^g)\) and \(\sin(\theta_j^g-\theta_i^g)\), respectively, mirroring the historical orientation encoding. Like \(\phi_e\), \(\phi_g\) separately embeds position and orientation, sums the embeddings, and refines the sum. The relative endpoint indicates whether the agents are likely to become spatially close, while these directional terms describe their prospective orientation relation. Together, they provide information that cannot be obtained reliably from current distance alone.

Before the final GCERT layer, the predicted-goal embedding is added to the intermediate node representation as shown in Eq.~\eqref{eq:goal-node-update}:
\begin{equation}
\label{eq:goal-node-update}
\tilde{\mathbf{h}}_{i}^{g}
=
\mathbf{h}_{i}^{(L-1)} + \phi_{n}(\hat{\mathbf{g}}_{i}).
\end{equation}
The updated node representation \(\tilde{\mathbf{h}}_{i}^{g}\) is used throughout the final layer, and \(\mathbf{e}_{ij}^{g}\) replaces the history-only edge feature. Thus, goal information affects the attention queries and keys, relation routing, relation-specific value transformations, and the subsequent state-expert transformation. Let \(\check{\mathbf h}_i^g=\operatorname{LN}_1(\tilde{\mathbf h}_i^g)\) denote the normalized input to the final interaction sublayer. The latent-relation probabilities are
\begin{equation}
\label{eq:goal-relation-router}
\boldsymbol{\rho}_{ij}^{g}
=
\operatorname{SoftMax}
\left(
f_{\mathrm{rel}}
\left[
\check{\mathbf{h}}_{i}^{g}
\mathbin{\Vert}\check{\mathbf{h}}_{j}^{g}
\mathbin{\Vert}\mathbf{e}_{ij}^{g}
\right]
\right)
\end{equation}
and the relation-specific values
\begin{equation}
\label{eq:goal-relation-expert}
\mathbf{v}_{ij}^{g,r}
=
\mathcal{T}_{r}
\left(
\left[
\check{\mathbf{h}}_{j}^{g}
\mathbin{\Vert}\mathbf{e}_{ij}^{g}
\right]
\right).
\end{equation}
The final-layer router and expert in Eqs.~\eqref{eq:goal-relation-router} and \eqref{eq:goal-relation-expert} use prospective motion to determine not only which neighboring agents are relevant, but also how edge-level relation experts transform the transmitted information and how node-level state experts transform the aggregated representation. This differs from using goals only as decoder conditions: goal information modifies both node and edge representations before the final interaction computation and consequently influences the complete final-layer information flow. After a final layer normalization, the interaction-enhanced features are projected into mode-specific representations for downstream decoding.

\subsection{Multimodal Decoder and Training Objective}

Following the multimodal Laplace decoding scheme of HiVT~\cite{zhou2022hivt}, the decoder combines the local motion feature \(\mathbf h_i\) and each mode-specific global feature \(\mathbf f_i^k\) to predict a trajectory location, a positive scale, and a mode logit:
\begin{equation}
\label{eq:decoder}
(\hat{\boldsymbol\mu}_{i,k},\mathbf b_{i,k},z_i^k)
=\mathcal D(\mathbf h_i,\mathbf f_i^k),\qquad
\pi_i^k=\operatorname{SoftMax}_k(z_i^k).
\end{equation}
The raw scale output $\mathbf{s}_{i,k}$ is transformed elementwise as $\mathbf{b}_{i,k}=\operatorname{ELU}(\mathbf{s}_{i,k})+1+b_{\min}$, with $b_{\min}=10^{-3}$, to ensure strictly positive Laplace scales. The negative log-likelihood additionally clamps scales to its numerical lower bound. All predicted trajectories and goal offsets use the agent-centric local frame. As in HiVT, regression trains the mode with the smallest cumulative displacement error over valid future steps:
\begin{equation}
\label{eq:best-mode}
k_i^*=\arg\min_k\sum_t m_{it}\|\hat{\boldsymbol\mu}_{i,k,t}-\mathbf y_i^t\|_2,
\end{equation}
where \(m_{it}\) indicates a valid future observation. The total objective is
\begin{equation}
\label{eq:training-objective}
\mathcal{L}=\mathcal{L}_{\mathrm{reg}}+\mathcal{L}_{\mathrm{cls}}+\lambda_g\mathcal{L}_{\mathrm{goal}}.
\end{equation}
For the selected mode, \(\mathcal{L}_{\mathrm{reg}}\) averages \(\log(2b)+|y-\hat\mu|/b\) over all valid agent--time--coordinate entries. For each agent with \(T_i=\sum_t m_{it}>0\), the detached classification target is
\begin{equation}
p_i^{*,k}=\operatorname{SoftMax}_k\!\left(-\frac{1}{T_i}\sum_t m_{it}\|\hat{\boldsymbol\mu}_{i,k,t}-\mathbf y_i^t\|_2\right).
\end{equation}
The classification loss is \(\mathcal{L}_{\mathrm{cls}}=-\operatorname{mean}_{i:T_i>0}\sum_k p_i^{*,k}\log\pi_i^k\). Let $\mathbf{g}_i^*=\mathbf{y}_i^{T_f}$ denote the ground-truth final displacement in the agent-centric local frame, and let $N_g=\sum_i m_{i,T_f}$ count agents with an observed final future position. The goal loss averages smooth $L_1$ errors over valid agents and both coordinates:
\begin{equation}
\label{eq:goal-loss}
\mathcal{L}_{\mathrm{goal}}=\frac{1}{2N_g}\sum_{i:m_{i,T_f}=1}\sum_{d=1}^{2}
\ell_{\mathrm{SL1}}(\hat{g}_{i,d}-g_{i,d}^*),
\end{equation}
where $\ell_{\mathrm{SL1}}(u)=u^2/2$ for $|u|<1$ and $|u|-1/2$ otherwise. We set $\mathcal{L}_{\mathrm{goal}}=0$ when $N_g=0$ or goal prediction is disabled.

\section{Experiments}
\subsection{Experiments Setup}

\subsubsection{Dataset}

The proposed framework is evaluated on the Argoverse motion forecasting benchmark \cite{chang2019argoverse}, which contains 323,557 urban driving sequences collected in Miami and Pittsburgh. Following the standard protocol, each sequence provides 2 seconds of historical observations, and the task is to predict the target agent's trajectory over the subsequent 3 seconds. The official training, validation, and test splits contain 205,942, 39,472, and 78,143 sequences, respectively.
\subsubsection{Evaluation Metrics}

To assess multimodal trajectory prediction performance, we adopt three standard metrics: minimum Average Displacement Error (\(\mathrm{minADE}_K\)), minimum Final Displacement Error (\(\mathrm{minFDE}_K\)), and Miss Rate (\(\mathrm{MR}\)).

In our evaluation implementation, the mode with the smallest final displacement error is selected from the \(K\) predictions. \(\mathrm{minFDE}_K\) is its final-step L2 error, and the reported \(\mathrm{minADE}_K\) is its average L2 error across future time steps; ADE is not minimized independently over modes. \(\mathrm{MR}\) is the proportion of samples for which none of the \(K\) predicted endpoints lies within 2 meters of the ground-truth endpoint.
\subsubsection{Implementation Details}

All experiments are implemented in PyTorch. AdamW is used as the optimizer with a weight decay of $1\times10^{-4}$. The initial learning rate is set to \(1\times10^{-3}\) and annealed to zero using a cosine schedule over 64 epochs, with a batch size of 128. The embedding dimension is \(D=128\), the number of attention heads is \(H=8\), the feed-forward hidden dimension is \(4D=512\), and the dropout rate is 0.1. The decomposition kernel has length \(k=13\); the temporal encoder and GCERT use 2 and 3 layers, respectively, with 5 experts in each router. The goal-loss weight is \(\lambda_g=0.2\). The temporal Transformer uses ReLU activations in its feed-forward sublayers.
Following existing prediction methods \cite{lan2024traj,zhou2022hivt,lan2025diffutory}, 2 seconds of historical trajectory observations are used as model inputs, and the number of predicted modes is set to 6. 

\subsection{Quantitative Analysis}

In this section, the proposed framework is quantitatively compared with recent state-of-the-art trajectory prediction methods on the Argoverse benchmark.

\begin{table}[htbp]
    \centering
    \footnotesize
    \setlength{\tabcolsep}{1.5pt}
    
    \caption{Quantitative comparison on the Argoverse validation set. Results are rounded to two decimal places; $\mathrm{MR}_6$ is reported as a ratio. The best results within each category are highlighted in \textbf{bold}, while the second-best results are underlined. The symbol ``*'' indicates that the map module of an originally map-based method is removed. Lower values indicate better performance ($\downarrow$).}
    
    \label{tab:quantitative_argoverse}

    \begin{tabular}{cccccc}
        \toprule
        \shortstack{Map\\Type} & Method & Pub. & \(\mathrm{minADE}_6\downarrow\) & \(\mathrm{minFDE}_6\downarrow\) & \(\mathrm{MR}_6\downarrow\) \\
        \midrule
        \multirow{11}{*}{\shortstack{Map-\\free}}
        & WIMP* \cite{khandelwal2020if} & arXiv 20' & 0.86 & 1.39 & 0.16 \\
        & LaneGCN* \cite{liang2020learning} & ECCV 20' & 0.79 & 1.29 & 0.17 \\
        & TPCN* \cite{ye2021tpcn} & CVPR 21' & 0.82 & 1.32 & 0.15 \\
        & STAM-P \cite{hou2024vehicle} & IoTJ 24' & 0.82 & 1.30 & 0.15 \\
        & CRAT-Pred \cite{schmidt2022crat} & ICRA 22' & 0.85 & 1.44 & 0.17 \\
        & HSTI \cite{luo2025hsti} & T-ITS 25' & 0.80 & 1.35 & 0.16 \\
        & HiVT-64* \cite{zhou2022hivt} & CVPR 22' & 0.77 & \underline{1.25} & \underline{0.14} \\
        & BSM \cite{gomez2023efficient} & T-ITS 24' & 0.81 & 1.56 & -- \\
        & MLB \cite{ren2025mlb} & IoTJ 25' & 0.77 & \underline{1.25} & \underline{0.14} \\
        & MFIG-Net \cite{wang2025mfig} & IoTJ 25' & \underline{0.76} & \underline{1.25} & \underline{0.14} \\
        & Ours & -- & \textbf{0.72} & \textbf{1.13} & \textbf{0.12} \\
        \midrule
        \multirow{12}{*}{\shortstack{Map-\\based}}
        & LaneGCN \cite{liang2020learning} & ECCV 20' & \textbf{0.71} & 1.08 & -- \\
        & mmTransformer \cite{liu2021multimodal} & CVPR 21' & \textbf{0.71} & 1.08 & \underline{0.10} \\
        & TPCN \cite{ye2021tpcn} & CVPR 21' & 0.82 & 1.32 & 0.11 \\
        & AutoBot-Ego \cite{girgis2021latent} & ICLR 22' & 0.73 & 1.10 & 0.12 \\
        & DSP \cite{zhang2022trajectory} & IROS 22' & \underline{0.72} & \textbf{0.99} & \textbf{0.09} \\
        & LTP \cite{wang2022ltp} & CVPR 22' & 0.78 & 1.07 & -- \\
        & MultiLane \cite{sierra2022multilane} & ITSC 22' & 0.84 & 1.42 & 0.11 \\
        & HGO \cite{mo2023map} & RAL 23' & 0.78 & 1.45 & -- \\
        & HLT-Ens \cite{lafage2025hierarchical} & WACV 25' & 0.78 & 1.24 & -- \\
        & I2T \cite{zhou2024i2t} & T-ITS 24' & 0.76 & 1.20 & 0.12 \\
        & UrbanTP \cite{zhang2024research} & T-ITS 24' & \textbf{0.71} & 1.14 & -- \\
        & MacFormer \cite{feng2023macformer} & RAL 23' & \textbf{0.71} & \underline{1.05} & \underline{0.10} \\
   \bottomrule
    \end{tabular}
\end{table}

Table~\ref{tab:quantitative_argoverse} compares prediction accuracy on Argoverse. The proposed framework achieves \(\mathrm{minADE}_6=0.72\), \(\mathrm{minFDE}_6=1.13\), and \(\mathrm{MR}_6=0.12\), obtaining the best results across all three metrics among the reported map-free approaches. Compared with MFIG-Net, the strongest competing method in \(\mathrm{minADE}_6\), our model reduces \(\mathrm{minADE}_6\) by 0.04 (5.3\%), \(\mathrm{minFDE}_6\) by 0.12 (9.6\%), and \(\mathrm{MR}_6\) by 0.02 (14.3\%). It also outperforms the map-free variants of HiVT-64 and MLB by 0.05 in \(\mathrm{minADE}_6\), 0.12 in \(\mathrm{minFDE}_6\), and 0.02 in \(\mathrm{MR}_6\).

The improvements are more pronounced relative to earlier map-free interaction models. In comparison with CRAT-Pred, the proposed method decreases \(\mathrm{minADE}_6\) from 0.85 to 0.72 and \(\mathrm{minFDE}_6\) from 1.44 to 1.13, corresponding to relative reductions of 15.3\% and 21.5\%, respectively; meanwhile, \(\mathrm{MR}_6\) is reduced from 0.17 to 0.12. Similar gains over STAM-P and HSTI show that the advantage is not confined to a single displacement metric. The simultaneous reductions in displacement error and miss rate indicate that the complete framework improves trajectory precision without sacrificing endpoint coverage.

These results establish the overall effectiveness of the complete framework. The contributions of trend--detail motion modeling, relation-aware message routing, and goal-conditioned interaction reasoning are examined separately in the following ablation study.

Map-based trajectory prediction approaches exploit lane geometry and road connectivity to constrain future trajectories, and therefore have access to stronger scene priors than the proposed method. Nevertheless, under the standard \(K=6\) setting, our map-free framework remains competitive with several representative map-based approaches. It improves upon HGO from 0.78/1.45 to 0.72/1.13 in \(\mathrm{minADE}_6/\mathrm{minFDE}_6\), and also surpasses I2T in both displacement metrics while attaining the same \(\mathrm{MR}_6\) of 0.12. Compared with TPCN, the proposed method substantially lowers both average and endpoint errors, although TPCN retains a marginally lower miss rate.

The proposed method does not outperform every map-based approach. DSP and MacFormer, for example, still benefit from explicit map constraints and obtain lower endpoint errors. This remaining gap is expected because lane topology directly restricts the space of feasible future paths, whereas a map-free predictor must infer such structure indirectly from short motion histories and agent interactions. Even so, the competitive results against multiple map-based baselines demonstrate that informative intra-agent motion modeling and prospective relation reasoning can recover a substantial portion of the predictive information otherwise supplied by HD maps.

\subsection{Ablation Study}
We conduct three groups of experiments on the Argoverse validation set to evaluate the core modeling components, goal-based interaction modeling, and the number of experts. The complete model achieves a $\mathrm{minADE}_6$ of 0.721, a $\mathrm{minFDE}_6$ of 1.128, and an $\mathrm{MR}_6$ of 11.90\%.

\subsubsection{Core Modeling Components}
Table~\ref{tab:ablation_components} evaluates feature decomposition, message MoE, and state MoE by removing each component individually while retaining the others. All three removals degrade every metric, indicating that none of these components is redundant in the evaluated architecture. Removing feature decomposition produces the largest increase in $\mathrm{minADE}_6$ and $\mathrm{MR}_6$, from 0.721 to 0.736 and from 11.90\% to 12.30\%, respectively. The result suggests that the subsequent interaction layers do not fully compensate for the loss of structured intra-agent motion encoding. Separating the smooth trend from local motion variations provides component-specific representations before temporal aggregation, which may help preserve both persistent motion tendencies and short-term deviations. This is particularly relevant to map-free prediction, where historical motion must provide cues that cannot be obtained from explicit lane geometry.

Disabling message or state MoE reduces the corresponding expert count to one ($R=1$ or $S=1$), leaving a single shared transformation without multi-expert routing. The two MoE ablations reveal different performance profiles. Removing message experts increases $\mathrm{minADE}_6$ and $\mathrm{minFDE}_6$ by 0.013 and 0.032, whereas removing state experts increases them by 0.010 and 0.028. Conversely, the miss-rate increase is slightly larger without state experts (0.30 percentage points) than without message experts (0.20 percentage points). These differences are consistent with the distinct positions of the two routing mechanisms: message experts construct relation-dependent information before aggregation, while state experts adapt the receiving agent's representation after aggregation. In particular, the degradation without message experts supports the motivation that selecting relevant neighbors alone is insufficient; the transformation of their messages also matters. The state-expert result further indicates that relation-aware messages do not eliminate the benefit of adaptive post-aggregation processing. Together, the ablations support retaining both stages, although the metric differences alone do not establish exclusive functional roles for either expert set.

\begin{table}[htbp]
    \centering
    \caption{Ablation of the core modeling components on the Argoverse validation set. Lower values are better.}
    \label{tab:ablation_components}
    \setlength{\tabcolsep}{3pt}
    \renewcommand{\arraystretch}{1.1}
    \resizebox{\columnwidth}{!}{
    \begin{tabular}{ccc ccc}
        \toprule
        \shortstack{Feature\\Decom.} & \shortstack{Message\\MoE} & \shortstack{State\\MoE}
        & $\mathrm{minADE}_6\downarrow$ & $\mathrm{minFDE}_6\downarrow$ & $\mathrm{MR}_6(\%)\downarrow$ \\
        \midrule
        \checkmark & \checkmark & \checkmark & \textbf{0.721} & \textbf{1.128} & \textbf{11.90} \\
                   & \checkmark & \checkmark & 0.736 & 1.160 & 12.30 \\
        \checkmark &            & \checkmark & 0.734 & 1.160 & 12.10 \\
        \checkmark & \checkmark &            & 0.731 & 1.156 & 12.20 \\
        \bottomrule
    \end{tabular}}
    \par\smallskip
    \footnotesize Note: $\checkmark$ indicates an enabled component. State MoE denotes the node-level FFN MoE. Goal prediction and goal-based edge updates are retained in all variants.
\end{table}

\subsubsection{Goal-Based Interaction Modeling}
Table~\ref{tab:ablation_goals} separates the contribution of goal prediction from the additional benefit of using predicted goals to update interaction edges. Relative to the variant with neither component, retaining goal prediction reduces $\mathrm{minFDE}_6$ from 1.163 to 1.151 and $\mathrm{MR}_6$ from 12.50\% to 12.10\%, while also slightly improving $\mathrm{minADE}_6$. This suggests that prospective motion information is useful even without explicitly incorporating it into pairwise edge geometry. However, this variant still falls short of the complete model, indicating that retaining goal prediction alone does not realize the full benefit of goal-based interaction modeling.

Adding goal-based edge updates lowers $\mathrm{minADE}_6$ by 0.007, $\mathrm{minFDE}_6$ by 0.023, and $\mathrm{MR}_6$ by 0.20 percentage points. The additional displacement-error reductions exceed those obtained by retaining goal prediction alone. This comparison highlights the importance of how prospective information enters the interaction computation: predicted goals can describe not only individual motion tendencies but also future relative configurations between agents. Encoding these configurations in the edges allows the final interaction layer to condition neighbor relevance and message construction on prospective pairwise relations, rather than relying only on historical relation features. Relative to the variant with neither component, the complete model reduces $\mathrm{minFDE}_6$ by approximately 3.0\% and $\mathrm{MR}_6$ by 0.60 percentage points. The consistent improvements support using predicted goals for relational modeling in addition to goal prediction itself.

\begin{table}[htbp]
    \centering
    \caption{Ablation of goal prediction and goal-based edge updates on the Argoverse validation set.}
    \label{tab:ablation_goals}
    \setlength{\tabcolsep}{4pt}
    \renewcommand{\arraystretch}{1.1}
    \resizebox{\columnwidth}{!}{
    \begin{tabular}{cc ccc}
        \toprule
        \shortstack{Goal\\Prediction} & \shortstack{Goal Edge\\Update}
        & $\mathrm{minADE}_6\downarrow$ & $\mathrm{minFDE}_6\downarrow$ & $\mathrm{MR}_6(\%)\downarrow$ \\
        \midrule
        \checkmark & \checkmark & \textbf{0.721} & \textbf{1.128} & \textbf{11.90} \\
        \checkmark &            & 0.728 & 1.151 & 12.10 \\
                   &            & 0.731 & 1.163 & 12.50 \\
        \bottomrule
    \end{tabular}}
\end{table}

\begin{figure*}[!t]
    \centering
    \includegraphics[width=\textwidth,height=0.36\textheight,keepaspectratio]{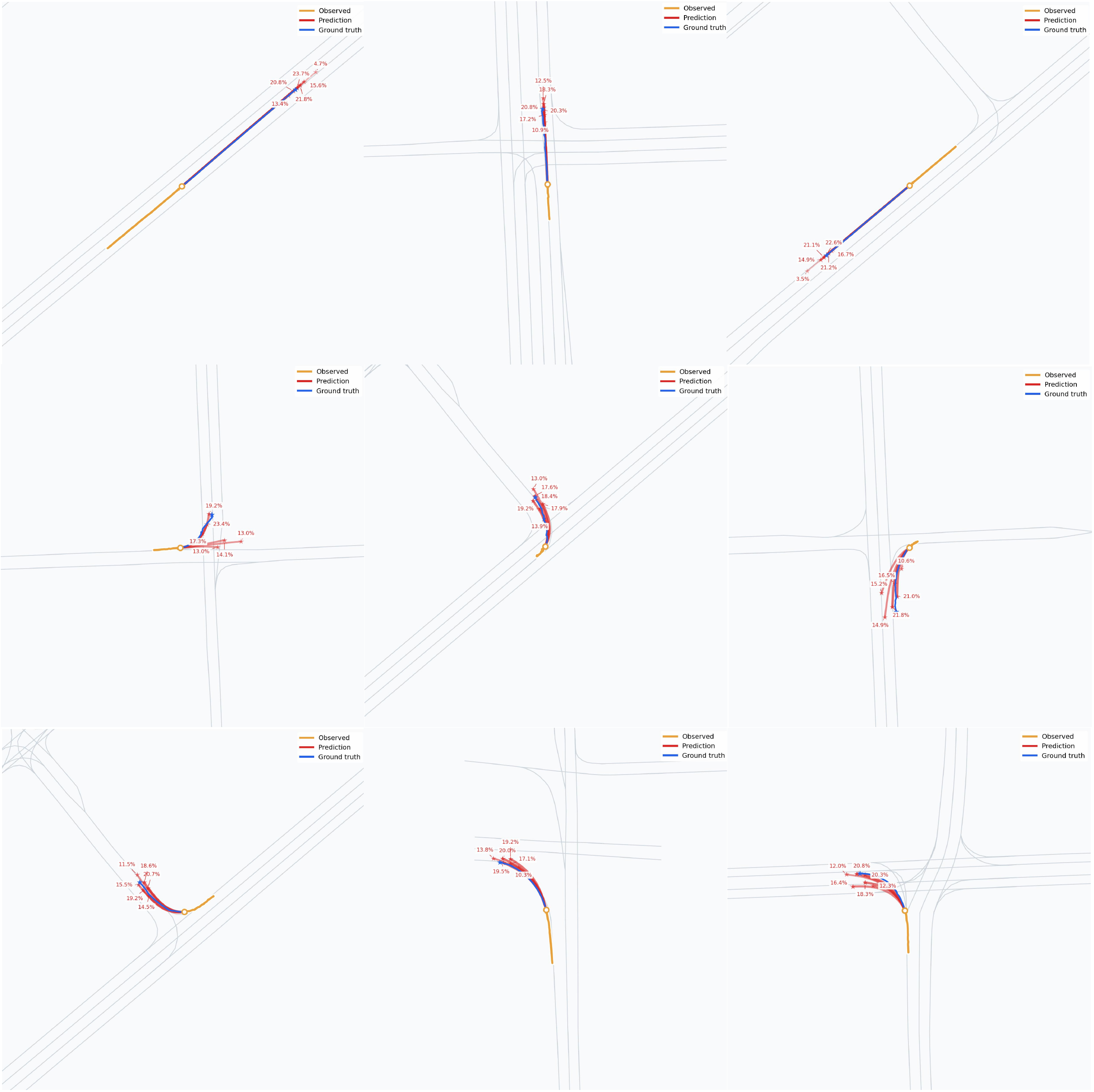}
    \caption{Qualitative trajectory predictions in representative driving scenarios. The first row shows straight-driving cases, and the second and third rows show turning cases. Orange, red, and blue curves denote observed trajectories, multimodal predictions, and ground-truth futures, respectively. Road geometry is shown for visual reference only; the model does not use HD maps as input.}
    \label{fig:vis}
\end{figure*}

\subsubsection{Number of Experts}
Table~\ref{tab:ablation_expert_counts} examines whether increasing the number of expert transformations consistently improves prediction. Starting from four experts in each MoE, increasing only the message or state expert count to five leaves $\mathrm{minADE}_6$ unchanged at 0.725, but lowers $\mathrm{minFDE}_6$ from 1.138 to 1.135 or 1.133, respectively. Using five experts in both stages further reduces the displacement metrics to 0.721 and 1.128. This pattern suggests that providing additional transformation choices at both message construction and state adaptation is more effective for displacement accuracy than expanding either stage alone within the tested four-to-five range.

The benefit does not continue as the expert count increases. With the other stage fixed at five experts, increasing either count to six worsens both displacement metrics; using six experts in both stages also fails to improve the results. Thus, the observed gains cannot be explained simply by a monotonic increase in expert capacity. A possible interpretation is that a moderate expert set provides sufficient transformation diversity, while further expansion introduces additional routing and optimization demands without useful gains on this dataset. The expert-count experiment does not by itself establish redundancy or overfitting, but it provides no empirical support for the larger configurations tested here.

The best configuration also depends on the metric: four message experts and five state experts yield the lowest $\mathrm{MR}_6$ of 11.80\%, whereas five experts in each stage yield the lowest $\mathrm{minADE}_6$ and $\mathrm{minFDE}_6$, with an $\mathrm{MR}_6$ of 11.90\%. This is not contradictory, since miss rate measures threshold exceedances rather than average displacement magnitude. We therefore select five experts in each MoE for the best displacement accuracy among the evaluated configurations, while acknowledging the small miss-rate advantage of the four/five setting.

\begin{table}[htbp]
    \centering
    \caption{Effect of the numbers of message and state experts on the Argoverse validation set. Bold indicates the best value in each metric.}
    \label{tab:ablation_expert_counts}
    \setlength{\tabcolsep}{4pt}
    \renewcommand{\arraystretch}{1.1}
    \resizebox{\columnwidth}{!}{
    \begin{tabular}{cc ccc}
        \toprule
        \shortstack{Message\\Experts} & \shortstack{State\\Experts}
        & $\mathrm{minADE}_6\downarrow$ & $\mathrm{minFDE}_6\downarrow$ & $\mathrm{MR}_6(\%)\downarrow$ \\
        \midrule
        4 & 4 & 0.725 & 1.138 & 12.00 \\
        4 & 5 & 0.725 & 1.133 & \textbf{11.80} \\
        5 & 4 & 0.725 & 1.135 & 11.90 \\
        5 & 5 & \textbf{0.721} & \textbf{1.128} & 11.90 \\
        5 & 6 & 0.727 & 1.145 & 12.00 \\
        6 & 5 & 0.725 & 1.137 & 12.00 \\
        6 & 6 & 0.727 & 1.146 & 12.00 \\
        \bottomrule
    \end{tabular}}
\end{table}

\subsection{Qualitative Analysis}

\subsubsection{Scenario-level Visualization}
Fig.~\ref{fig:vis} presents qualitative prediction results for nine representative scenarios from the Argoverse validation set. The first row shows relatively simple straight-driving cases, while the second and third rows show turning maneuvers. Observed trajectories, predicted trajectories, and ground-truth futures are shown in orange, red, and blue, respectively. The road geometry is displayed only for visual reference and is not used as input to the model.

In the straight-driving cases, the predicted trajectory sets contain hypotheses that closely follow the ground truth, capturing the overall direction and extent of future motion. In particular, the middle and right examples in the first row exhibit strong agreement between the predicted paths and the ground-truth trajectories. The left example also retains a turning alternative alongside straight-going predictions, illustrating that the multimodal output can represent different possible continuations near an intersection.

The turning cases require the model to anticipate changes in heading and curved future paths rather than simply extrapolate the observed direction. Across the second and third rows, the predictions include trajectories that closely match the ground-truth turning direction and overall curvature. Although some hypotheses differ in turning radius or endpoint position, and several scenes retain straight-going alternatives, the predicted sets generally cover the realized maneuvers. These examples qualitatively support the ability of the proposed framework to capture both persistent straight-line motion and nonlinear turning behavior without explicit map inputs, complementing the quantitative evaluation across diverse driving scenarios.

\begin{figure*}[!t]
    \centering
    \includegraphics[width=\textwidth]{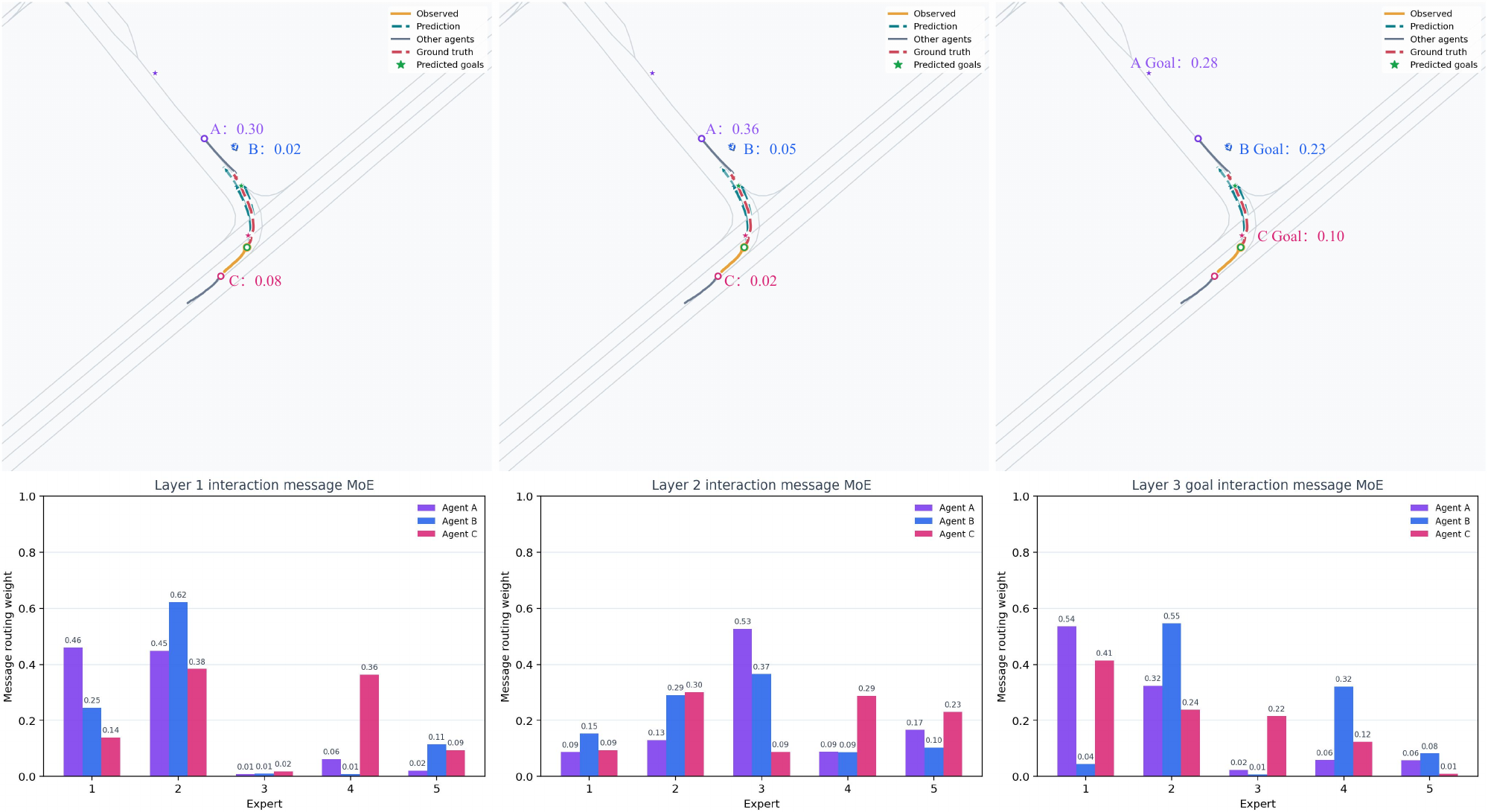}
    \caption{Visualization of relation-aware message routing across three inter-agent learning layers. The top row shows interaction scores in the same scene, with predicted goals used to update the edges before Layer 3. Agents A, B, and C are the three highest-scoring neighbors in Layer 3: a preceding vehicle traveling in the same direction, a stationary vehicle, and a following vehicle, respectively. Colored stars denote the corresponding agents' predicted goals. The bottom row shows their message-expert routing weights in each layer. Experts are independently parameterized in each layer; expert indices do not imply shared parameters or corresponding semantics across layers.}
    \label{fig:message_expert_vis}
\end{figure*}

\subsubsection{Visualization of Relation-Aware Message Routing}
Fig.~\ref{fig:message_expert_vis} presents the interaction scores and message-expert routing weights across three inter-agent learning layers in a representative scene. The top row shows the interaction scores, with the third layer using edges updated from predicted goals, while the bottom row shows the corresponding Message MoE weights. We select Agents A, B, and C, the three highest-scoring neighbors in the goal-conditioned layer, and examine their scores and routing distributions throughout the network. Agent A is a preceding vehicle traveling in the same direction as the focal agent, Agent B is stationary, and Agent C follows behind the focal agent.

In the first two layers, the focal agent primarily attends to Agent A, whose interaction scores are 0.30 and 0.36, respectively. Agents B and C receive substantially lower scores: 0.02 and 0.08 in Layer 1, and 0.05 and 0.02 in Layer 2. This emphasis on Agent A is consistent with the car-following relation in the scene. Within Layer 1, Agent A's message is routed mainly through Experts 1 and 2, with weights of 0.46 and 0.45; within Layer 2, Expert 3 has the largest weight of 0.53. These layer-specific routing preferences may reflect the use of the respective experts to model car-following interactions. Since experts are independently parameterized in each layer, their indices are interpreted only within that layer, without assuming semantic correspondence across layers.

In the goal-conditioned third layer, Agent A retains a high interaction score of 0.28, while Agent B's score increases from 0.05 in Layer 2 to 0.23. The focal agent thus assigns relatively high scores to both A and B. A possible explanation for the increased score of B is the proximity between the focal agent's predicted goal and B's predicted goal, which may strengthen their association in goal-based interaction reasoning. Agent C's score also increases to 0.10 but remains below those of A and B. The resulting attention distribution therefore differs from that in the preceding layers, rather than exhibiting a uniform increase across neighbors.

Despite their similar interaction scores in Layer 3, Agents A and B exhibit distinct expert distributions. Agent A's message is mainly processed by Experts 1 and 2, with weights of 0.54 and 0.32, whereas Agent B's message is mainly processed by Experts 2 and 4, with weights of 0.55 and 0.32. These different mixtures may reflect distinct treatments of the car-following relation with A and the interaction with stationary vehicle B. The two messages share an expert component but place different emphasis on the remaining transformations. Agent C exhibits another soft mixture, with weights of 0.41, 0.24, and 0.22 on Experts 1, 2, and 3, respectively.

These observations illustrate the complementary roles of interaction scoring and relation-aware message routing. Interaction scores regulate the relative contributions of neighboring agents, while expert weights determine the mixture of transformations used to construct each message. Predicted-goal proximity offers a possible explanation for the greater attention to B in the final layer, while the distinct routing distributions of A and B show that similarly relevant neighbors need not be processed in the same way. This example supports relation-dependent message construction through soft expert mixtures, without establishing a fixed semantic category for any individual expert.

\subsubsection{Visualization of Maneuver-Dependent State Expert Routing}
\begin{figure*}[!t]
    \centering
    \includegraphics[width=\textwidth]{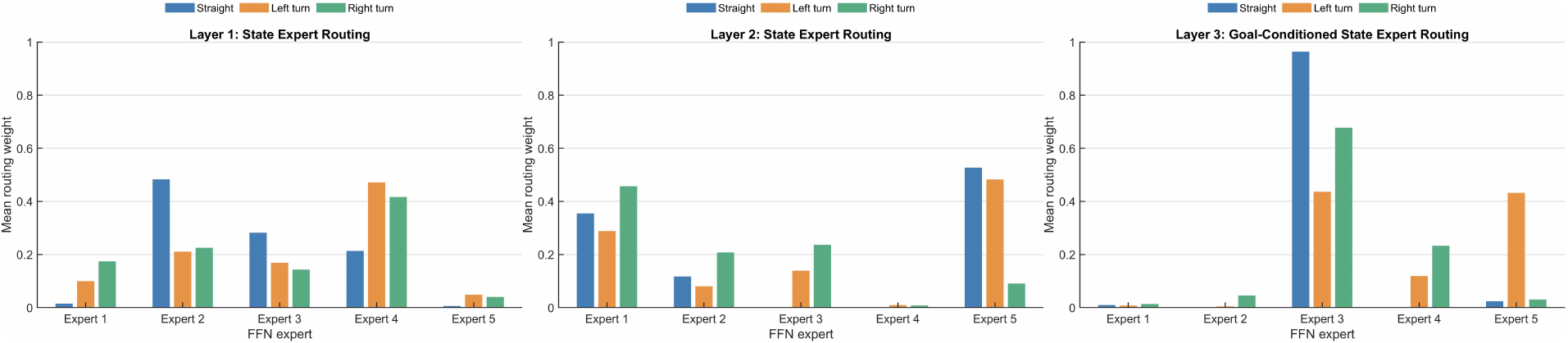}
    \caption{Maneuver-dependent state-expert routing across three interaction layers. Bars show the mean soft routing weight of each expert within straight-driving, left-turning, and right-turning scenes. The first two layers model historical interactions, while the third operates on goal-updated representations. Expert parameters are independent across layers; identical expert indices do not imply corresponding semantics.}
    \label{fig:state_expert_vis}
\end{figure*}

To examine whether state-expert routing varies across motion behaviors, we group the evaluated scenes using the bearing angle $\alpha=\operatorname{atan2}(y_y,y_x)$ of the focal agent's last valid ground-truth future displacement $(y_x,y_y)$ in its current local coordinate frame. Angles above $+30^{\circ}$ indicate left turns, angles below $-30^{\circ}$ indicate right turns, and the remaining angles indicate straight driving. Scenes with fewer than two valid future steps or an endpoint displacement norm below $10^{-6}$~m are assigned to straight driving. These labels are used only for post-hoc routing analysis. We compute the mean routing weight of each expert within each category. Fig.~\ref{fig:state_expert_vis} shows the resulting distributions across the three interaction layers. Averaging within each category allows routing preferences to be compared despite the unequal numbers of scenes.

Maneuver-dependent routing preferences are already observable in the historical interaction layers. In Layer 1, Expert 2 receives a higher mean weight for straight driving (0.483) than for left and right turns (0.211 and 0.225), whereas Expert 4 receives higher weights for both turning categories. Layer 2 exhibits an additional distinction between turning directions: Expert 5 receives substantial weights for straight driving and left turns (0.527 and 0.483), but a much lower weight for right turns (0.091), for which Expert 1 has the largest mean weight (0.457). These patterns indicate that the state router responds differently to representations associated with different maneuvers.

In the goal-conditioned third layer, straight-driving scenes place most of their routing weight on Expert 3 (0.964). Left-turning scenes allocate comparable mean weights to Experts 3 and 5 (0.436 and 0.432), while right-turning scenes primarily weight Experts 3 and 4 (0.677 and 0.233). Expert 3 therefore remains prominent across all three categories, with turning scenes assigning additional weight to different experts. For straight-driving scenes, the category-averaged routing distribution becomes increasingly concentrated with depth: its largest expert weight rises from 0.483 in Layer 1 to 0.527 in Layer 2 and 0.964 in Layer 3. Turning categories do not exhibit the same consistent trend and retain more distributed average routing weights in the final layer. Thus, the observed concentration trend depends on motion behavior rather than applying uniformly across all maneuver categories.

These observations support maneuver-dependent state routing rather than a one-to-one correspondence between experts and maneuver classes. They complement the relation-aware message-routing visualization by showing adaptive expert weighting at the receiving-agent state level. The category averages describe aggregate routing preferences, rather than the mixture used by every individual scene. Moreover, the cross-layer comparison alone does not isolate the effect of goal conditioning, since the layers differ in both their parameters and input representations.

\subsection{Real-Vehicle Experiment}
\begin{figure}[!t]
    \centering
    \includegraphics[width=\columnwidth]{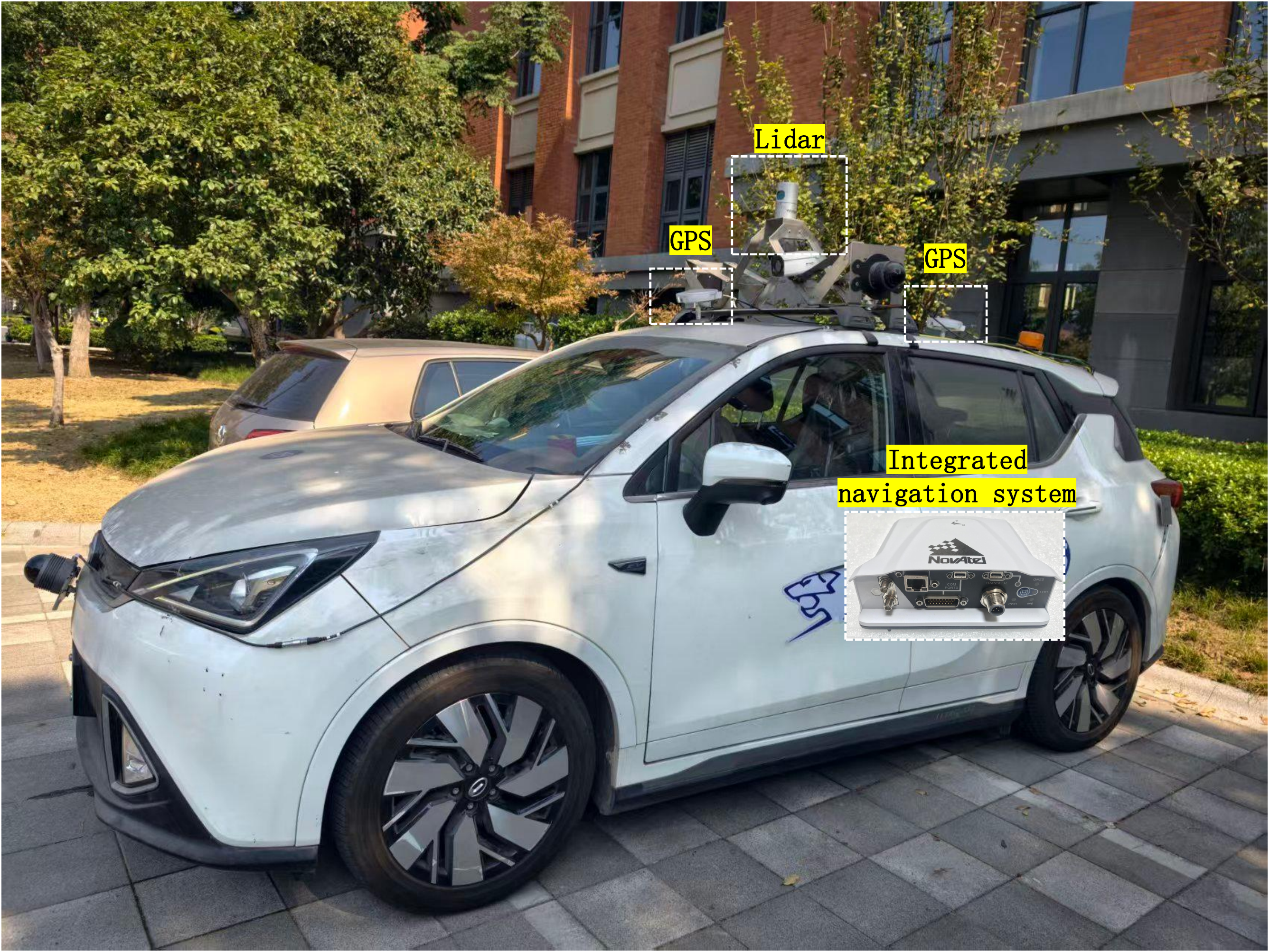}
    \caption{Experimental vehicle and its perception and localization equipment. The LiDAR provides point-cloud observations, while GPS and the integrated navigation system provide ego-vehicle positioning for trajectory coordinate alignment.}
    \label{fig:real_vehicle}
\end{figure}

To examine the proposed model with trajectories obtained from vehicle-mounted sensors, we use recordings collected on campus roads. The experimental vehicle, shown in Fig.~\ref{fig:real_vehicle}, carries a LiDAR, GPS antennas, and an integrated navigation system. Detected and tracked traffic participants are transformed into a common coordinate frame using the ego-vehicle pose. The predictor receives historical position sequences of the target and available neighboring agents; the point cloud and road layout are not supplied as model inputs. We apply the Argoverse-trained model without campus-specific fine-tuning and use approximately 2~s of history sampled at 10~Hz to generate six candidate trajectories over the following 3~s. Subsequent tracked positions serve as the reference future for these examples.

\begin{figure*}[!t]
    \centering
    \begin{minipage}[t]{0.49\textwidth}
        \centering
        \includegraphics[width=\linewidth]{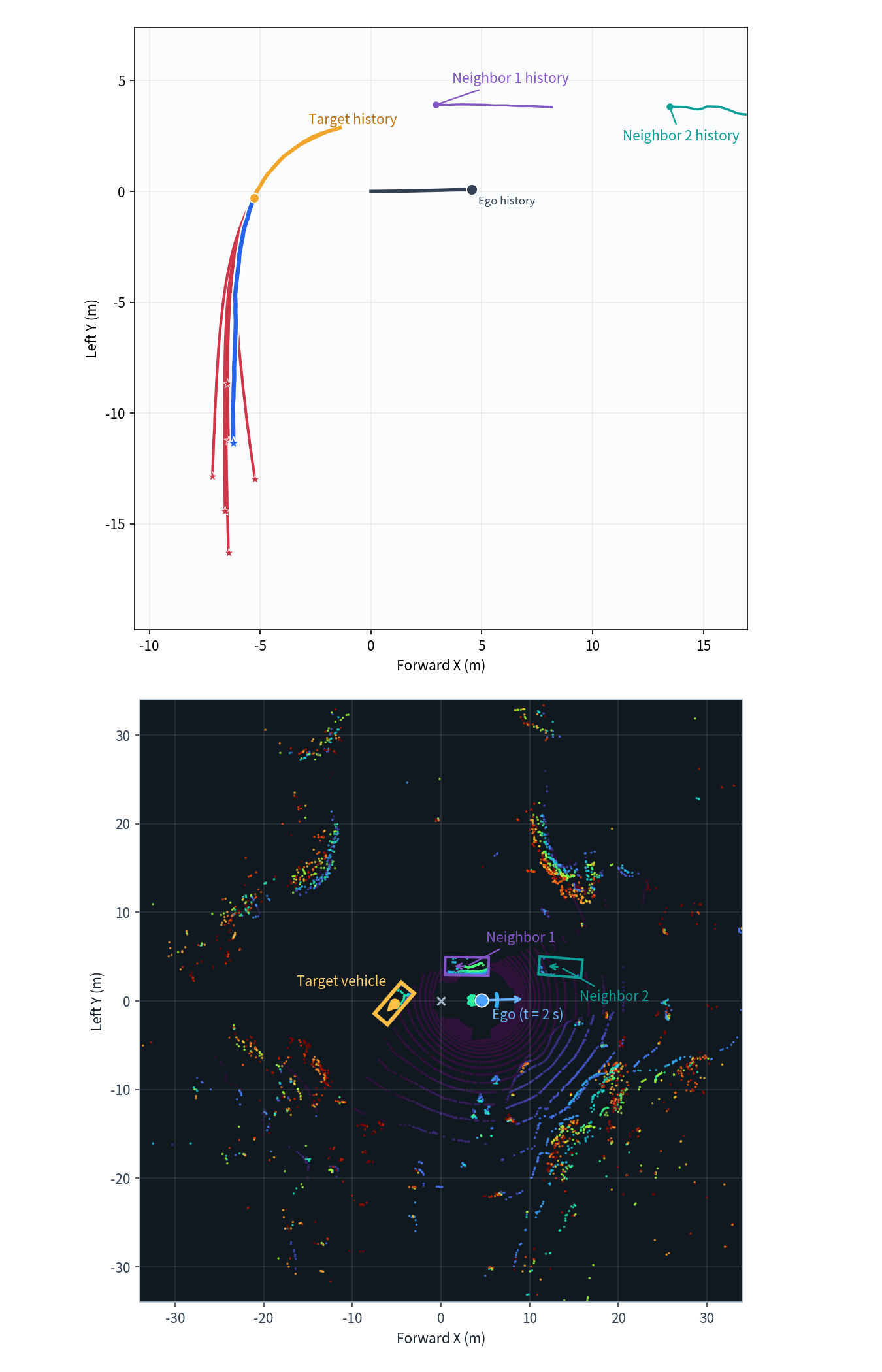}
        \par\smallskip\footnotesize (a) Curved-turn case.
    \end{minipage}\hfill
    \begin{minipage}[t]{0.49\textwidth}
        \centering
        \includegraphics[width=\linewidth]{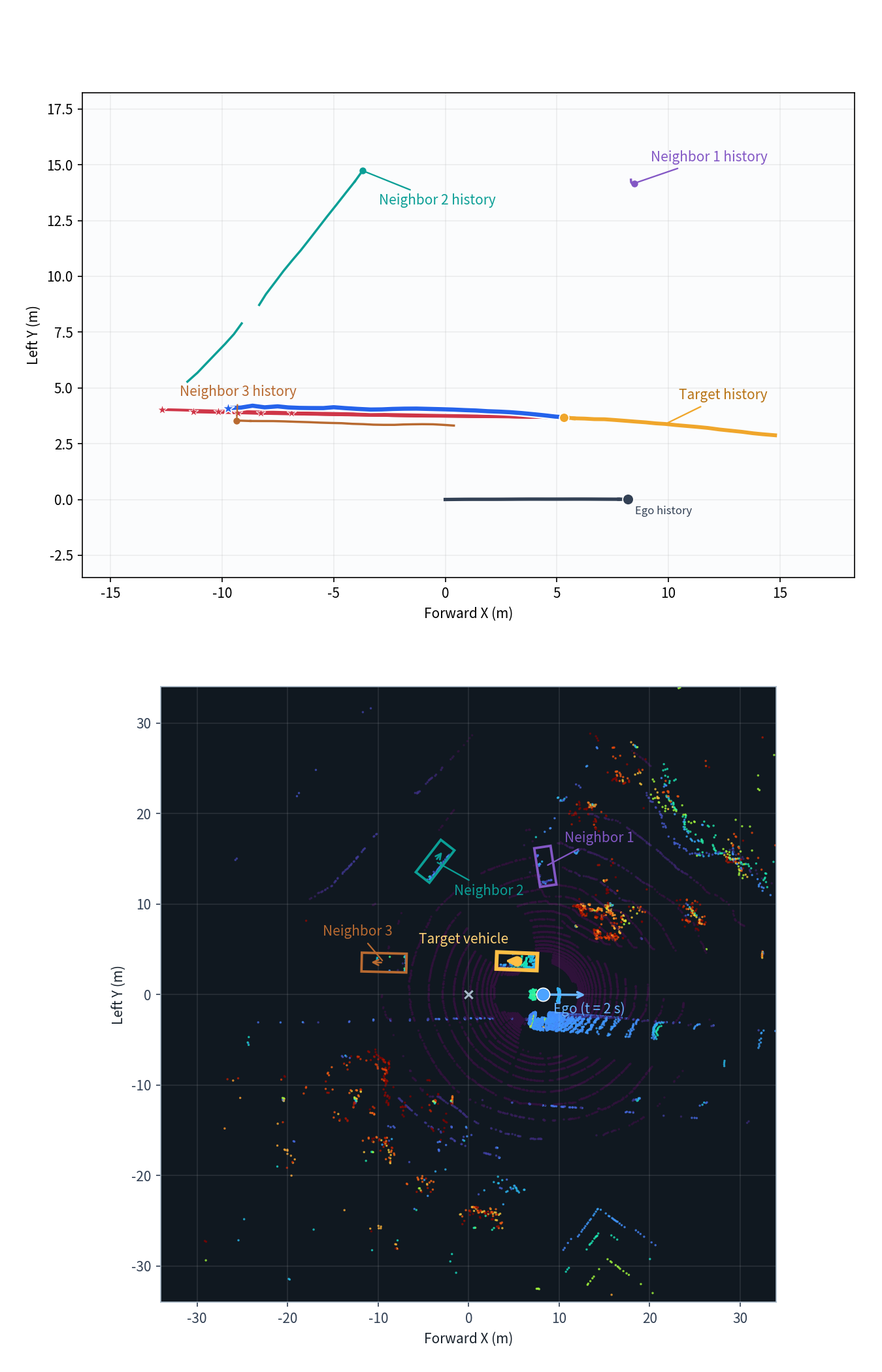}
        \par\smallskip\footnotesize (b) Multi-vehicle near-straight case.
    \end{minipage}
    \caption{Selected real-vehicle predictions. Each upper panel shows target history (orange), ego history (dark gray), available neighbor histories (other colors), six predicted candidates (red), and the subsequently tracked reference future (blue); stars mark future endpoints. Each lower panel shows the corresponding LiDAR point cloud and tracked objects at the final observation time. The six-mode minFDE values are 0.22~m in (a) and 0.48~m in (b).}
    \label{fig:real_vehicle_predictions}
\end{figure*}

Fig.~\ref{fig:real_vehicle_predictions} shows two selected cases with different target motions. In the curved-turn case, the target history already changes direction, and the candidate futures continue along the observed turning direction while varying in lateral position and travel distance. One candidate endpoint lies close to the subsequently tracked endpoint, with a six-mode minimum final displacement error (minFDE) of 0.22~m. In the multi-vehicle near-straight case, the target and several neighbors are tracked simultaneously. The candidate futures follow the target's main direction of travel and retain modest differences in endpoint and lateral offset; the six-mode minFDE is 0.48~m. The lower panels connect each trajectory plot to the point-cloud frame and tracked objects at the end of the observation window. These selected examples illustrate candidate coverage under real sensor observations; they do not provide an aggregate accuracy estimate for the campus recordings.

\FloatBarrier

\section{Conclusion}
In summary, this study proposes a map-free trajectory prediction framework centered on a Goal-Conditioned Expert-Routed Transformer. A lightweight intra-agent front-end separates projected histories into a smooth trend and a local detail residual, applies component-specific residual projections, and summarizes their recombined evolution through causal temporal self-attention. GCERT combines edge-level relation experts with node-level state experts, while predicted motion goals update the node and edge representations entering the final interaction layer. The resulting goal-conditioned attention, message routing, and state transformation enable prospective inter-agent reasoning without explicit road topology. Experiments on the Argoverse benchmark demonstrate competitive prediction accuracy, while ablation studies verify the complementary contributions of structured motion encoding, expert-routed message generation, and goal-conditioned interaction reasoning. The selected real-vehicle examples additionally show candidate trajectories obtained from tracked sensor observations without campus-specific fine-tuning or HD-map inputs. A limitation of the map-free setting is that, in complex scenes that rely heavily on road-marking guidance, such as large unmarked roundabouts or diverging ramps without informative historical agent trajectories, history-only reasoning may be more prone to off-road predictions than methods with strong map constraints. Future work will investigate adaptive smoothing mechanisms and mode-dependent goal relations to further improve motion representation and multimodal interaction modeling.

 \bibliographystyle{IEEEtran} 
\bibliography{IEEEabrv,myref}

% Generated by IEEEtran.bst, version: 1.14 (2015/08/26)
\begin{thebibliography}{10}
\providecommand{\url}[1]{#1}
\csname url@samestyle\endcsname
\providecommand{\newblock}{\relax}
\providecommand{\bibinfo}[2]{#2}
\providecommand{\BIBentrySTDinterwordspacing}{\spaceskip=0pt\relax}
\providecommand{\BIBentryALTinterwordstretchfactor}{4}
\providecommand{\BIBentryALTinterwordspacing}{\spaceskip=\fontdimen2\font plus
\BIBentryALTinterwordstretchfactor\fontdimen3\font minus \fontdimen4\font\relax}
\providecommand{\BIBforeignlanguage}[2]{{%
\expandafter\ifx\csname l@#1\endcsname\relax
\typeout{** WARNING: IEEEtran.bst: No hyphenation pattern has been}%
\typeout{** loaded for the language `#1'. Using the pattern for}%
\typeout{** the default language instead.}%
\else
\language=\csname l@#1\endcsname
\fi
#2}}
\providecommand{\BIBdecl}{\relax}
\BIBdecl

\bibitem{gao2020vectornet}
J.~Gao, C.~Sun, H.~Zhao, Y.~Shen, D.~Anguelov, C.~Li, and C.~Schmid, ``Vector{N}et: Encoding hd maps and agent dynamics from vectorized representation,'' in \emph{Proceedings of the IEEE/CVF conference on Computer Vision and Pattern Recognition}, 2020, pp. 11\,525--11\,533.

\bibitem{liang2020learning}
M.~Liang, B.~Yang, R.~Hu, Y.~Chen, R.~Liao, S.~Feng, and R.~Urtasun, ``Learning lane graph representations for motion forecasting,'' in \emph{European Conference on Computer Vision}, 2020, pp. 541--556.

\bibitem{jia2023hdgt}
X.~Jia, P.~Wu, L.~Chen, Y.~Liu, H.~Li, and J.~Yan, ``H{DGT}: Heterogeneous driving graph transformer for multi-agent trajectory prediction via scene encoding,'' \emph{IEEE transactions on pattern analysis and machine intelligence}, vol.~45, no.~11, pp. 13\,860--13\,875, 2023.

\bibitem{xiong2025fine}
W.~Xiong, J.~Chen, and Z.~Qi, ``Fine-grained behavior and lane constraints guided trajectory prediction method,'' \emph{arXiv preprint arXiv:2503.21477}, 2025.

\bibitem{gomez2023efficient}
C.~G{\'o}mez-Hu{\'e}lamo, M.~V. Conde, R.~Barea, M.~Oca{\~n}a, and L.~M. Bergasa, ``Efficient baselines for motion prediction in autonomous driving,'' \emph{IEEE Transactions on Intelligent Transportation Systems}, vol.~25, no.~5, pp. 4192--4205, 2023.

\bibitem{xin2025multi}
G.~Xin, D.~Chu, L.~Lu, Z.~Deng, Y.~Lu, and X.~Wu, ``Multi-agent trajectory prediction with difficulty-guided feature enhancement network,'' \emph{IEEE Robotics and Automation Letters}, 2025.

\bibitem{liu2024laformer}
M.~Liu, H.~Cheng, L.~Chen, H.~Broszio, J.~Li, R.~Zhao, M.~Sester, and M.~Y. Yang, ``Laformer: Trajectory prediction for autonomous driving with lane-aware scene constraints,'' in \emph{Proceedings of the IEEE/CVF conference on Computer Vision and Pattern Recognition}, 2024, pp. 2039--2049.

\bibitem{liu2021multimodal}
Y.~Liu, J.~Zhang, L.~Fang, Q.~Jiang, and B.~Zhou, ``Multimodal motion prediction with stacked transformers,'' in \emph{Proceedings of the IEEE/CVF conference on Computer Vision and Pattern Recognition}, 2021, pp. 7577--7586.

\bibitem{feng2023macformer}
C.~Feng, H.~Zhou, H.~Lin, Z.~Zhang, Z.~Xu, C.~Zhang, B.~Zhou, and S.~Shen, ``Macformer: Map-agent coupled transformer for real-time and robust trajectory prediction,'' \emph{IEEE Robotics and Automation Letters}, vol.~8, no.~10, pp. 6795--6802, 2023.

\bibitem{ren2025mlb}
Y.~Ren, L.~Liu, Z.~Lan, Z.~Cui, and H.~Yu, ``Mlb-traj: Map-free trajectory prediction with local behavior query for autonomous driving,'' \emph{IEEE Internet of Things Journal}, 2025.

\bibitem{luo2025hsti}
X.~Luo, S.~Fu, B.~Gao, Y.~Zhao, H.~Tan, and Z.~Song, ``Hsti: A light hierarchical spatial-temporal interaction model for map-free trajectory prediction,'' \emph{IEEE Transactions on Intelligent Transportation Systems}, 2025.

\bibitem{schmidt2022crat}
J.~Schmidt, J.~Jordan, F.~Gritschneder, and K.~Dietmayer, ``Crat-pred: Vehicle trajectory prediction with crystal graph convolutional neural networks and multi-head self-attention,'' in \emph{2022 International Conference on Robotics and Automation}, 2022, pp. 7799--7805.

\bibitem{zhou2022hivt}
Z.~Zhou, L.~Ye, J.~Wang, K.~Wu, and K.~Lu, ``Hi{VT}: Hierarchical vector transformer for multi-agent motion prediction,'' in \emph{Proceedings of the IEEE/CVF conference on Computer Vision and Pattern Recognition}, 2022, pp. 8823--8833.

\bibitem{shi2022motion}
S.~Shi, L.~Jiang, D.~Dai, and B.~Schiele, ``Motion transformer with global intention localization and local movement refinement,'' \emph{Advances in Neural Information Processing Systems}, vol.~35, pp. 6531--6543, 2022.

\bibitem{lan2024traj}
Z.~Lan, L.~Liu, B.~Fan, Y.~Lv, Y.~Ren, and Z.~Cui, ``Traj-llm: A new exploration for empowering trajectory prediction with pre-trained large language models,'' \emph{IEEE Transactions on Intelligent Vehicles}, 2024.

\bibitem{hou2024vehicle}
Y.~Hou, X.~Zhang, H.~Zhang, X.~Cao, Z.~Lu, and X.~Yuan, ``A vehicle trajectory prediction model for map-free scenes using the spatiotemporal attentional mechanism,'' \emph{IEEE Internet of Things Journal}, vol.~12, no.~9, pp. 11\,372--11\,382, 2024.

\bibitem{xiang2023map}
J.~Xiang, Z.~Nan, Z.~Song, J.~Huang, and L.~Li, ``Map-free trajectory prediction in traffic with multi-level spatial-temporal modeling,'' \emph{IEEE Transactions on Intelligent Vehicles}, vol.~9, no.~2, pp. 3258--3270, 2023.

\bibitem{wang2025mfig}
W.~Wang, Z.~Ma, X.~Liu, S.~Yang, Y.~Yang, and H.~Xia, ``Mfig-net: Map-free multi-agent trajectory prediction for signalized intersection scenarios,'' \emph{IEEE Internet of Things Journal}, 2025.

\bibitem{chung2014empirical}
J.~Chung, C.~Gulcehre, K.~Cho, and Y.~Bengio, ``Empirical evaluation of gated recurrent neural networks on sequence modeling. arxiv 2014,'' \emph{arXiv preprint arXiv:1412.3555}, vol. 1412, 2014.

\bibitem{hochreiter1997long}
S.~Hochreiter and J.~Schmidhuber, ``Long short-term memory,'' \emph{Neural computation}, vol.~9, no.~8, pp. 1735--1780, 1997.

\bibitem{xu2023context}
P.~Xu, J.-B. Hayet, and I.~Karamouzas, ``Context-aware timewise vaes for real-time vehicle trajectory prediction,'' \emph{IEEE Robotics and Automation Letters}, vol.~8, no.~9, pp. 5440--5447, 2023.

\bibitem{vaswani2017attention}
A.~Vaswani, N.~Shazeer, N.~Parmar, J.~Uszkoreit, L.~Jones, A.~N. Gomez, {\L}.~Kaiser, and I.~Polosukhin, ``Attention is all you need,'' \emph{Advances in neural information processing systems}, vol.~30, 2017.

\bibitem{gu2024mamba}
A.~Gu and T.~Dao, ``Mamba: Linear-time sequence modeling with selective state spaces,'' in \emph{First conference on language modeling}, 2024.

\bibitem{bian2024multi}
Y.~Bian, X.~Ju, J.~Li, Z.~Xu, D.~Cheng, and Q.~Xu, ``Multi-patch prediction: adapting language models for time series representation learning,'' in \emph{Forty-first International Conference on Machine Learning}, 2024.

\bibitem{chen2024pathformer}
P.~Chen, Y.~Zhang, Y.~Cheng, Y.~Shu, Y.~Wang, Q.~Wen, B.~Yang, and C.~Guo, ``Pathformer: Multi-scale transformers with adaptive pathways for time series forecasting,'' \emph{arXiv preprint arXiv:2402.05956}, 2024.

\bibitem{zhang2023flight}
Z.~Zhang, D.~Guo, S.~Zhou, J.~Zhang, and Y.~Lin, ``Flight trajectory prediction enabled by time-frequency wavelet transform,'' \emph{Nature Communications}, vol.~14, no.~1, p. 5258, 2023.

\bibitem{zhang2024g2ltraj}
Z.~Zhang, Z.~Hua, M.~Chen, W.~Lu, B.~Lin, D.~Cai, and W.~Wang, ``G2ltraj: A global-to-local generation approach for trajectory prediction,'' \emph{arXiv preprint arXiv:2404.19330}, 2024.

\bibitem{yuan2021agentformer}
Y.~Yuan, X.~Weng, Y.~Ou, and K.~M. Kitani, ``Agentformer: Agent-aware transformers for socio-temporal multi-agent forecasting,'' in \emph{Proceedings of the IEEE/CVF International Conference on Computer Vision}, 2021, pp. 9813--9823.

\bibitem{girgis2021latent}
R.~Girgis, F.~Golemo, F.~Codevilla, M.~Weiss, J.~A. D'Souza, S.~E. Kahou, F.~Heide, and C.~Pal, ``Latent variable sequential set transformers for joint multi-agent motion prediction,'' \emph{arXiv preprint arXiv:2104.00563}, 2021.

\bibitem{li2024efficient}
L.~Li, X.~Wang, J.~Lian, J.~Zhao, and J.~Hu, ``Efficient vehicle trajectory prediction with goal lane segments and dual-stream cross attention,'' \emph{IEEE Transactions on Intelligent Transportation Systems}, 2024.

\bibitem{gulzar2025gc}
M.~Gulzar, Y.~Muhammad, and N.~Muhammad, ``Gc-gat: Multimodal vehicular trajectory prediction using graph goal conditioning and cross-context attention,'' \emph{IEEE Robotics and Automation Letters}, 2025.

\bibitem{shazeer2017outrageously}
N.~Shazeer, A.~Mirhoseini, K.~Maziarz, A.~Davis, Q.~Le, G.~Hinton, and J.~Dean, ``Outrageously large neural networks: The sparsely-gated mixture-of-experts layer,'' \emph{arXiv preprint arXiv:1701.06538}, 2017.

\bibitem{chang2019argoverse}
M.-F. Chang, J.~Lambert, P.~Sangkloy, J.~Singh, S.~Bak, A.~Hartnett, D.~Wang, P.~Carr, S.~Lucey, D.~Ramanan \emph{et~al.}, ``Argoverse: 3d tracking and forecasting with rich maps,'' in \emph{Proceedings of the IEEE/CVF conference on Computer Vision and Pattern Recognition}, 2019, pp. 8748--8757.

\bibitem{lan2025diffutory}
Z.~Lan, L.~Liu, Y.~Ren, Z.~Cui, and H.~Yu, ``Diffutory: A future feature and mode association augmented diffusion model for trajectory prediction in autonomous vehicles,'' \emph{IEEE Transactions on Intelligent Transportation Systems}, 2025.

\bibitem{khandelwal2020if}
S.~Khandelwal, W.~Qi, J.~Singh, A.~Hartnett, and D.~Ramanan, ``What-if motion prediction for autonomous driving,'' \emph{arXiv preprint arXiv:2008.10587}, 2020.

\bibitem{ye2021tpcn}
M.~Ye, T.~Cao, and Q.~Chen, ``Tpcn: Temporal point cloud networks for motion forecasting,'' in \emph{Proceedings of the IEEE/CVF Conference on Computer Vision and Pattern Recognition}, 2021, pp. 11\,318--11\,327.

\bibitem{zhang2022trajectory}
L.~Zhang, P.~Li, J.~Chen, and S.~Shen, ``Trajectory prediction with graph-based dual-scale context fusion,'' in \emph{2022 IEEE/RSJ International Conference on Intelligent Robots and Systems (IROS)}.\hskip 1em plus 0.5em minus 0.4em\relax IEEE, 2022, pp. 11\,374--11\,381.

\bibitem{wang2022ltp}
J.~Wang, T.~Ye, Z.~Gu, and J.~Chen, ``Ltp: Lane-based trajectory prediction for autonomous driving,'' in \emph{Proceedings of the IEEE/CVF Conference on Computer Vision and Pattern Recognition}, 2022, pp. 17\,134--17\,142.

\bibitem{sierra2022multilane}
D.~Sierra-Gonzalez, A.~Paigwar, O.~Erkent, and C.~Laugier, ``Multilane: Lane intention prediction and sensible lane-oriented trajectory forecasting on centerline graphs,'' in \emph{2022 IEEE 25th International Conference on Intelligent Transportation Systems (ITSC)}.\hskip 1em plus 0.5em minus 0.4em\relax IEEE, 2022, pp. 3657--3664.

\bibitem{mo2023map}
X.~Mo, Y.~Xing, H.~Liu, and C.~Lv, ``Map-adaptive multimodal trajectory prediction using hierarchical graph neural networks,'' \emph{IEEE Robotics and Automation Letters}, vol.~8, no.~6, pp. 3685--3692, 2023.

\bibitem{lafage2025hierarchical}
A.~Lafage, M.~Barbier, G.~Franchi, and D.~Filliat, ``Hierarchical light transformer ensembles for multimodal trajectory forecasting,'' in \emph{Proceedings of the Winter Conference on Applications of Computer Vision}, 2025, pp. 1682--1691.

\bibitem{zhou2024i2t}
Y.~Zhou, Z.~Wang, N.~Ning, Z.~Jin, N.~Lu, and X.~Shen, ``I2t: from intention decoupling to vehicular trajectory prediction based on prioriformer networks,'' \emph{IEEE Transactions on Intelligent Transportation Systems}, vol.~25, no.~8, pp. 9411--9426, 2024.

\bibitem{zhang2024research}
S.~Zhang, R.~Bai, R.~He, Z.~Meng, Y.~Chang, Y.~Zhi, and N.~Sun, ``Research on vehicle trajectory prediction methods in urban main road scenarios,'' \emph{IEEE Transactions on Intelligent Transportation Systems}, vol.~25, no.~11, pp. 16\,392--16\,408, 2024.

\end{thebibliography}

\end{document}